\documentclass[10pt,twocolumn,letterpaper]{article}

\usepackage[review]{cvpr}      

\renewcommand{\paragraph}[1]{\vspace{1pt}\noindent\textbf{#1}}

\usepackage{tcolorbox}
\tcbuselibrary{skins, breakable}

\newtcolorbox[auto counter, number within=section]{PromptBox}[3][]{
  sharp corners,
  boxrule=0.6pt,
  fonttitle=\bfseries,
  colback=#2!5,       
  colframe=#2!50,     
  title={#3}, 
  left=6pt,
  right=6pt,
  top=6pt,
  bottom=6pt,
  before skip=10pt,
  after skip=10pt,
  parbox=false,
  #1
}

\definecolor{cvprblue}{rgb}{0.21,0.49,0.74}
\usepackage[pagebackref,breaklinks,colorlinks,allcolors=cvprblue]{hyperref}

\def\confName{CVPR}
\def\confYear{2026}

\title{HarmoVid: Relightful Video Portrait Harmonization}

\author{
	Jun Myeong Choi$^{1}$\quad Jae Shin Yoon$^{2}$\quad Luchao Qi$^{1}$\quad Roni Sengupta$^{1}$\quad Joon-Young Lee$^{2}$\\
    \vspace{-5pt}\\
	$^{1}$University of North Carolina at Chapel Hill\qquad $^{2}$Adobe Research
	\\
	\url{https://chedgekorea.github.io/HarmoVid}
}

\usepackage{tikz}

\begin{document}

\twocolumn[{%
\renewcommand\twocolumn[1][]{#1}%
\maketitle
\includegraphics[width=\linewidth]{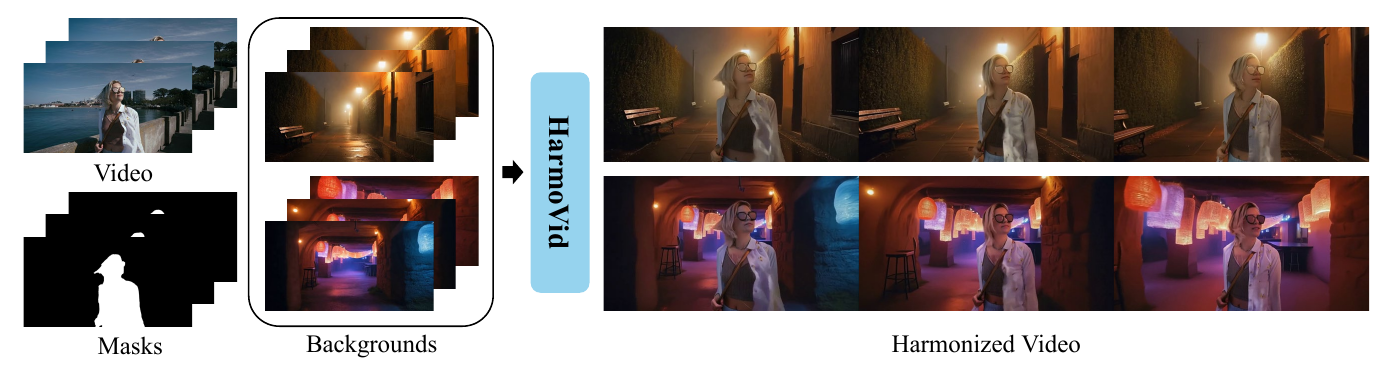}
\vspace{-2.5em}
\captionof{figure}{
We present HarmoVid, a relightable video harmonization framework that aligns lighting between foreground and background. Given a foreground video with masks and a target background video, HarmoVid leverages a video diffusion model to directly synthesize a harmonized output. It produces realistic, temporally consistent results that accurately match foreground color, intensity, and shadows to the background while preserving both foreground and background content.}
\label{fig:teaser}
\vspace{1em}
}]

\maketitle
\begin{abstract}
We present a method for harmonizing the lighting of a foreground video to match a target background scene, adjusting shadows, color tone, and illumination intensity (relightful harmonization). 
Unlike images, acquiring labeled data for videos, where identical motions are recorded under different lighting conditions, is practically infeasible and non-scalable.
While one way to create such paired data is to apply existing image-based harmonization models frame by frame to a video, the resulting outputs often suffer from significant temporal jitters.
We overcome this problem by introducing a novel lighting deflickering model that can stabilize the global and local lighting flickering artifacts.
Our video diffusion model learns from these upgraded deflickered data with a volume of real and synthetic videos to generate high-quality video harmonization results.
We further propose an asymmetric alpha mask conditioning technique to learn the clean boundaries from real videos.
Experiments demonstrate that our model achieves strong temporal coherence, naturalness, cleaner boundaries, and physically meaningful lighting behavior, while maintaining strong relighting expressiveness compared to prior image-based and video-based harmonization methods.

\end{abstract}
    
\section{Introduction}
\label{sec:intro}

Video harmonization, the task of adjusting a foreground object's appearance to seamlessly blend with a new background, has attracted increasing attention for realistic video composition in various applications such as filmmaking, video editing, and augmented reality. Despite significant progress in image harmonization, video harmonization remains a challenging task due to the additional complexity of the temporal dimension. Image harmonization techniques~\cite{ren2024relightful,imgharm,cha2025text2relight,zhang2025iclight} applied independently to each frame of a video often lead to temporal inconsistencies in lighting, causing noticeable flickering and making these approaches unsuitable for video harmonization. 

The primary obstacle in directly extending image harmonization models to videos is the scarcity of paired training data; it is practically impossible to capture an identical performance under varied lighting and background conditions, especially when recording human subjects whose motions, expressions, and poses cannot be precisely reproduced across multiple takes. Although synthetic data can be generated at scale, it still fails to represent subtle lighting–appearance interactions and realistic temporal behaviors observed in real scenes, which limits its effectiveness as a substitute.
This lack of high-quality, real-world data severely constrains the generalization capabilities of existing image harmonization models, particularly for harmonizing unconstrained, in-the-wild videos.

Existing approaches~\cite{zhang2025iclight,fang2025relightvid,zhang2025lumisculpt,cha2025text2relight,ren2024relightful,imgharm} have attempted to circumvent the data scarcity problem but often introduce their own limitations. 
Methods that extend frame-by-frame image harmonization techniques to video~\cite{video_relit1, video_relit2} often fail to maintain temporal coherence, resulting in distracting flickering artifacts. 
Others that rely on synthetic datasets~\cite{hou2021highfidelityfacerelighting,zhou2019deep,Song_2021} struggle to generalize to the complex lighting dynamics and natural variations of real-world scenes. Although recent generative networks offer more flexibility~\cite{lumen, liu2025tclight, zhou2025lightavid, zhang2025lumisculpt, fang2025relightvid}, they often inadvertently alter the core characteristics of the input, \textit{i.e.,} inadvertently alter the foreground or background contents, a problem known as identity shift. Furthermore, most methods~\cite{ zhou2025lightavid, fang2025relightvid, zhang2025iclight} are highly sensitive to the quality of the input foreground mask, leading to unnatural boundary artifacts when using imperfect masks, which is common in practical workflows. These limitations highlight the key properties that a `good' video harmonization model must satisfy: (a) preserve the identity and texture of the foreground and the background of the input video, (b) ensure temporal consistency, (c) exhibit robustness to imperfect masks, and (d) demonstrate high-fidelity expressiveness in lighting across diverse scenes. To date, no single framework has comprehensively addressed all these requirements.

\begin{figure}[t]
  \centering
  \includegraphics[width=\linewidth]{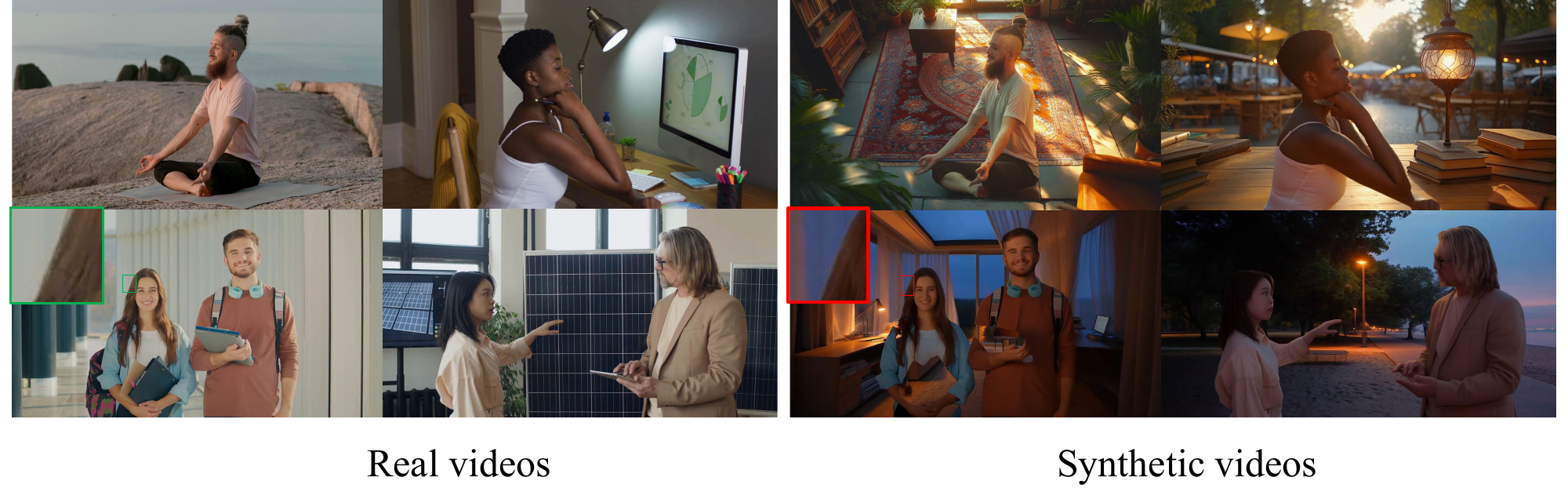}
  \caption{Samples from our paired video harmonization dataset. Real videos serve as high-quality ground truth, offering strong supervision for temporal coherence, physical plausibility, natural appearance, and clean boundaries. Synthetic videos, while highly scalable and cover a wide range of expressive and dynamic lighting conditions, inherently lack full realism and temporal fidelity. Our video harmonization model, HarmoVid, leverages the complementary strengths of both real and synthetic data to achieve robust and high-quality harmonization.}
  \vspace{-1.5em}
  \label{fig:data}
\end{figure}

To address these multifaceted challenges, we introduce a framework that achieves robust and temporally coherent video harmonization without requiring real paired data. Our approach is centered on a two-stage training strategy that combines data refinement with a robust harmonization model. We first composite real foreground videos onto various background videos, and generate pseudo-synthetic training pairs by applying an off-the-shelf image harmonization model to individual frames. Recognizing that this process introduces temporal flickering, we then train a lighting de-flickering network to refine these sequences, creating a high-quality and temporally consistent dataset. By jointly learning with real videos and refined synthetic videos, our video harmonization model, HarmoVid, could combine the strength of the videos from each domain (\textit{e.g.,} physical plausibility from real videos, and lighting expressiveness from synthetic videos as shown in Fig.~\ref{fig:data}).

We further improve the harmonization results on the boundary by asymmetric conditioning of the foreground mask: 
we use binary mask conditioning when learning from synthetic videos, while pseudo-alpha maps from real videos. By learning from real videos with such pseudo-alpha maps, HarmoVid could produce smooth foreground-background transitions, robust to imperfect segmentation during compositing and harmonizing.

We conducted extensive experiments on challenging real-world video datasets to evaluate our video harmonizaiton model. The results demonstrate that our method significantly outperforms state-of-the-art approaches in both quantitative metrics and qualitative assessments. Notably, HarmoVid produces videos with superior temporal stability, better identity preservation, cleaner boundary blending, and naturalness.

\begin{figure*}[h]
  \centering
  \includegraphics[width=\linewidth]{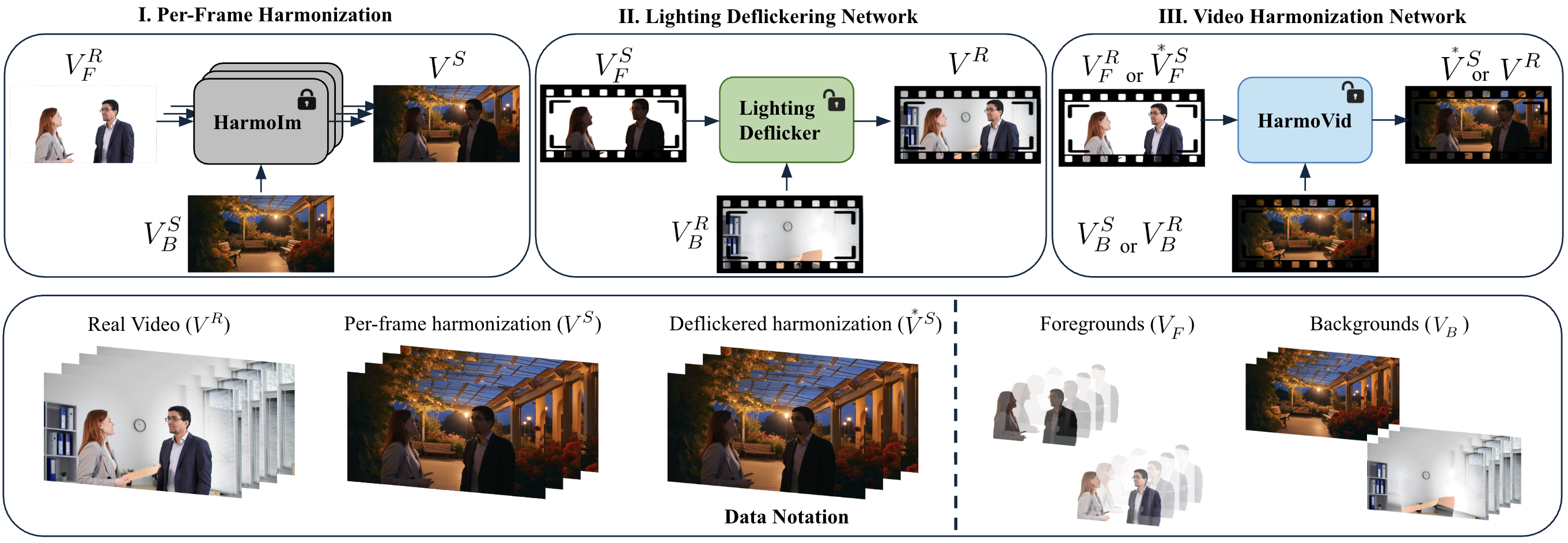}
  \vspace{-5mm}
  \caption{Our three-step pipeline for data synthesis and training for video harmonization. In the first state, we apply off-the-shelf image-based harmonization model to a video frame by frame to reconstruct the intermediate synthetic data. In the second stage, we upgrade the quality of the synthetic data by applying our lighting deflickering model to the synthetic videos. In the third step, HarmoVid jointly learns the real and synthetic data to generate temporally coherent, realistic, and expressive video harmonization results.  }
  \vspace{-0.5em}
  \label{fig:pipeline}
\end{figure*}

\section{Related Work}
\label{sec:rel_work}
\subsection{Image Relighting and Harmonization}

Image relighting aims to modify the illumination of a scene while preserving its underlying geometry and appearance. Most recent approaches are based on deep learning, capturing high-level relationships between scene structure and lighting and enabling controllable adjustments.

Deep learning–based relighting can be categorized into several subgroups. 3D-aware relighting approaches~\cite{3dr1,3dr2,3dr3,3dr4,3dr5,jin2024neural} utilize neural 3D representations to jointly model geometry and illumination, allowing complex lighting effects including viewpoint-dependent shadows and reflections. Inverse rendering-based methods~\cite{irb1,irb2,irb3,ponglertnapakorn2023difareli} explicitly decompose an image into reflectance, shading, and illumination components, providing physically interpretable results, though they often struggle to generalize to complex real-world scenes due to simplified assumptions about materials and lighting. Image-based relighting~\cite{ibr1,ibr2,ibr4,ibr5,scribblelight,ibr6,ibr7} with explicit lighting control leverages input images while providing additional cues such as light direction or intensity, enabling controllable and realistic relighting without requiring full 3D reconstruction.

In parallel, image harmonization focuses on adjusting the appearance of composite or inserted regions to match the lighting, tone, and color characteristics of the surrounding background \cite{ren2024relightful,imgharm}. Recent diffusion-based methods further enhance realism and robustness using large-scale synthetic or augmented datasets and enable controllable adjustments via image- or text-guided cues \cite{cha2025text2relight,zhang2025iclight}.

\subsection{Video Relighting and Harmonization}

Video relighting extends image relighting to sequences of frames, introducing the challenge of maintaining temporal consistency while ensuring spatially coherent illumination. Early works~\cite{video_relit1, video_relit2} used optical flow or temporal regularization to stabilize results across frames, whereas more recent approaches employ neural rendering, 3D-aware representations~\cite{video_3d1,video_3d2,video_3d3, video_3d4,video_3d_5,video_3d_6,video_3d_7,video_3d_8}—such as NeRF-based models and 3D Gaussian representations—or video diffusion models~\cite{video_diffusion1, video_diffusion2, video_diffusion3, video_diffusion4, qi2025over} to jointly model lighting, geometry, and temporal dynamics. While those methods allow controlling the lighting properties of the videos, they do not support the harmonization that automatically matches foreground and background lighting distribution.

Video harmonization shares the objective of maintaining consistent lighting, tone, and shading over time, particularly in composite or edited videos, while supporting the automatic lighting matching process. Recent methods further explore user-controllable or text-guided video harmonization. Some approaches \cite{lumen, liu2025tclight} enable text-to-video harmonization using video diffusion models. Light-A-Video~\cite{zhou2025lightavid} converts image-based harmonization models into video harmonization frameworks, whereas LumiSculpt\cite{zhang2025lumisculpt} employ light maps and light trajectories to allow user-directed manipulation of lighting within a single video, with LightCtrl\cite{anonymous2025lightctrl} additionally supporting user-controlled background lighting. However, all of these methods rely primarily on text prompts and cannot explicitly condition on the input background, effectively relighting the background along with the foreground. In contrast, RelightVid \cite{fang2025relightvid} leverages in-the-wild videos together with 3D-rendered data to achieve temporally consistent relighting under precise HDR illumination and explicitly conditions on the background. Nevertheless, when applied to real videos, it does not use paired data; instead, it employs foreground augmentation to adapt the model for real-world harmonization, which limits its ability to fully capture realistic lighting interactions and complex foreground–background relationships.
\section{Method}
\label{sec:method}

Our goal is to harmonize a given foreground video with a new background video by adjusting shadows, color tones, and illumination intensity in a visually coherent manner (\cref{fig:teaser}). The primary challenge lies in the scarcity of temporally consistent paired training data. To overcome this, we construct a large-scale pseudo-paired video dataset with diverse lighting conditions. Another challenge is the domain gap between real and synthetic videos. While real videos exhibit natural lighting and shadows, they lack sufficient diversity for learning expressive relighting effects. In contrast, synthetic videos can incorporate varied lighting conditions but often suffer from temporal artifacts.

To address these challenges, we employ a two-stage process for dataset generation: (1) per-frame image harmonization (Sec.~\ref{sec:perframe}) and (2) lighting deflickering (Sec.~\ref{sec:deflicker}). After constructing the dataset, we train a video harmonization model, HarmoVid, that learns a bidirectional mapping between real and synthetic domains (Sec.~\ref{sec:harmonization}). Fig.~\ref{fig:pipeline} provides an overview of the proposed framework.

\subsection{Expressive Paired Video Dataset Generation}
\label{sec:perframe}
We first describe the process of generating an expressive paired video dataset. To create data containing diverse lighting effects, we prepare real foreground videos $V_F^R$ containing target objects and synthetically generated background videos $V_B^S$. Foreground masks $V^M$ are obtained using Grounded-SAM-2~\cite{ren2024grounded} with the prompt ``the main character (human)." The background videos can be generated by variants of many video generators such as~\cite{videogen1,videogen2,videogen3} and are designed to include various lighting conditions. Similar to Text2Relight~\cite{cha2025text2relight}, we employ a range of lighting-related text prompts to control illumination characteristics.

Next, the foreground and background videos are composited, and a pretrained image harmonization model, HarmoIm~\cite{ren2024relightful}, is applied to each frame individually to adapt lighting and color tone accordingly. The process can be formulated as:
\begin{equation}
    V^S = \text{HarmoIm}(V_F^R, V_B^S, V^M),
\end{equation}
where $\text{HarmoIm}$ denotes the pretrained image harmonization model~\cite{ren2024relightful}.

Although this stage enables the construction of a large-scale pseudo-paired dataset $\{(V^S, V^R)\}$, the independent per-frame processing inevitably introduces temporal inconsistencies such as flickering and abrupt illumination transitions.
While one could apply existing video deflickering models~\cite{lei2023deflicker} to mitigate such issues, these methods are generally limited to stabilizing global frame flicker and struggle to handle local lighting variations (e.g., rapid local shadow changes between consecutive frames).
To resolve this limitation, we develop a lighting deflickering module that enhances temporal coherence across both global and local lighting components.

\begin{figure}[t]
  \centering
  \includegraphics[width=\linewidth]{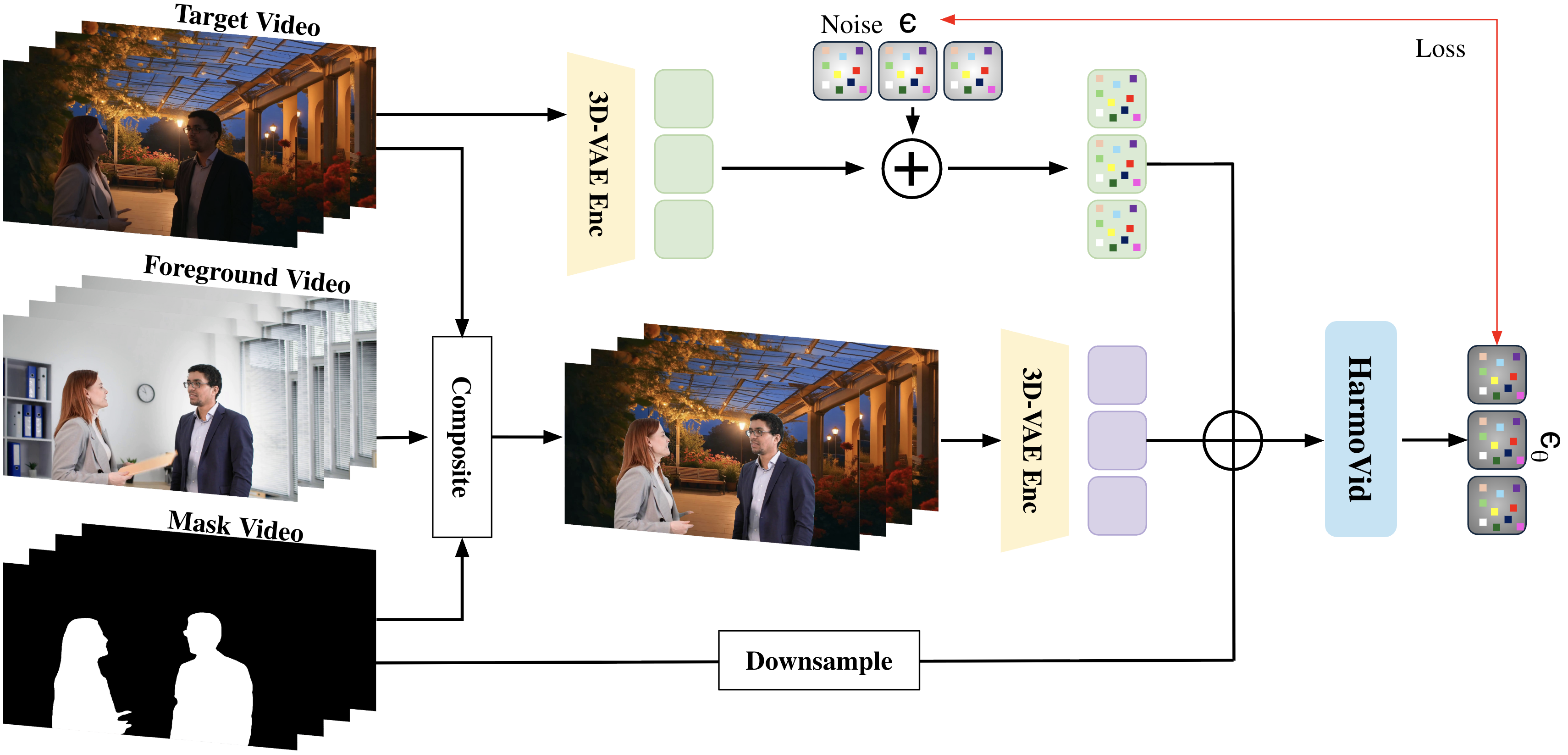}
  \caption{Our relightful video harmonization framework. HarmoVid is 3D transformer-based architecture that takes the input triplets of the latent space of the foregrounds with composited target backgrounds and noisy targets, and foreground masks; and predicts the noises that generates clean harmonized videos.}
  \label{fig:model_architecture}
\end{figure}

\subsection{Lighting Deflickering Network}
\label{sec:deflicker}

Our deflickering network is designed to stabilize temporal lighting coherence, addressing global color transitions and local shadow changes while maintaining spatial fidelity.  
To capture spatio-temporal dependencies effectively, we adopt a 3D latent diffusion transformer based on CogVideoX~\cite{yang2024cogvideox}, which extends the DiT architecture to model spatial and temporal dimensions jointly.

Our lighting deflickering video diffusion model learns to predict the noises given an input flickering video and the corresponding foreground masks, as illustrated in Fig.~\ref{fig:pipeline}:
\begin{equation}
  \mathcal{L}_\text{deflicker} = \mathbb{E}_{t, \epsilon}
  \left[ \left\| \epsilon - \epsilon_\theta(z^I, z^T_t, V^M, t) \right\|_2^2 \right],
\end{equation}
where $z^I$ and $z^T$ denote the latent representations of the synthetic composite $V_\text{comp}^S$ and the real target video $V^R$, respectively. The representations are encoded with a 3D-VAE encoder $\mathcal{E}(\cdot)$:
\begin{equation}
    z^I = \mathcal{E}(V_\text{comp}^S), \quad
    z^T = \mathcal{E}(V^R).
\end{equation}
Gaussian noise $\epsilon \sim \mathcal{N}(0,1)$ is added to $z^T$ according to diffusion timestep $t$. The synthetic composite $V_\text{comp}^S$ is produced by compositing the real background $V_B^R$ and the synthetic foreground $V_F^S$ using the foreground mask $V^M$:
\begin{equation}
    V_\text{comp}^S = \mathcal{C}(V_F^S, V_B^R, V^M),
\end{equation}
where $\mathcal{C}$ is the element-wise composite operation using the mask $V^M$.

After training this model, the frame-wise harmonized videos (generated in Sec.~\ref{sec:perframe}) $V^S$ and their corresponding masks $V^M$ are provided as inputs to the network. The model then produces temporally coherent videos $\hat{V}^S$, which exhibit reduced flickering and temporally coherent transitions under the background scenes.

\subsection{Video Harmonization Network}
\label{sec:harmonization}
We train our final video harmonization model, HarmoVid, using the paired video data constructed in the previous stage. The goal of HarmoVid is to adapt the lighting distribution of foregrounds to novel backgrounds, ensuring consistency in color and illumination while preserving temporal stability. As illustrated in Fig.~\ref{fig:model_architecture}, the foreground videos are first composited with the background videos using the triplet $\{V_F, V_B, V^M\}$, then the model produces harmonized results.

HarmoVid adopts the same 3D latent diffusion architecture as the deflickering network. It learns to predict noise in the latent space in order to reconstruct temporally coherent and spatially consistent harmonized videos:

\begin{equation}
  \mathcal{L}_\text{harm} = \mathbb{E}_{t, \epsilon}
  \left[ \left\| \epsilon - \epsilon_\theta(z^I, z^B, z^T_t, V^M, t) \right\|_2^2 \right],\label{objective:vidharmo}
\end{equation}
where $z^F$ and $z^B$ are the latent representations of the composited foreground and background, respectively, and $z^T$ is the noisy latent space of the ground-truth target videos. The associated latent representations are obtained by:
\begin{equation}
    z^I = \mathcal{E}(\mathcal{C}_\text{in}), \quad
    z^B = \mathcal{E}(V_B), \quad
    z^T = \mathcal{E}(V_\text{target}), 
\end{equation}
where $\mathcal{E}$ is the 3DVAE encoder~\cite{yang2024cogvideox}.

Depending on how the training data are organized within the dataset, HarmoVid learns from two distinct training paths. The first path takes a real foreground and a synthetically generated background as input to produce a harmonized synthetic video (Real $\rightarrow$ Synthetic). The second path employs processed foreground—adjusted to align with a new background condition—together with its corresponding real background to reconstruct a harmonized real video (Synthetic $\rightarrow$ Real). This dual-path training strategy enables the model to effectively bridge the distribution gap between real and synthetic domains, thereby enhancing its robustness and generalization capability.

\paragraph{Real $\rightarrow$ Synthetic path.}  
This training path takes real foregrounds $V_F^R$ composited with synthetic backgrounds $\hat{V}_B^S$ and binary mask $V^M$, to generate the synthetic harmonization video $\hat{V}^S$. Learning from this real-to-synthetic path is crucial for preserving the expressive relighting and shadow representation capabilities originally captured by the image-based harmonization model. 

\paragraph{Synthetic $\rightarrow$ Real path.}  
In this path, the foreground from a synthetic video $\hat{V_F^S}$ is composited with a real background $V_B^R$, where the foreground region is first removed using inpainting~\cite{Telea2004ImageIT}. A pseudo alpha mask $V^{\tilde{\alpha}}$ is used as a foreground mask during compositing instead of binary masks $V^M$. The pseudo alpha mask smoothly decays the foreground boundaries. Empirically, using the alpha mask in this synthetic-to-real path is critical for achieving natural and high-quality harmonization around object edges, as real videos provide perfect ground-truth blending supervision. In addition, this synthetic-to-real mapping enables the model to learn temporal coherence and physically consistent adjustments of shadows and illumination, inherited from real videos.

\begin{table*}[t]
  \caption{We compare HarmoVid with state-of-the-art image and video harmonization techniques to quantitatively evaluate harmonization quality (matching foreground and background lighting) and temporal consistency, along with qualitative evaluation with a user study.}
  \vspace{-0.5em}
  \label{tab:main_comparison}
  \centering
  \resizebox{\linewidth}{!}{%
  \begin{tabular}{lccccccccc}
    \toprule
    & \multicolumn{4}{c}{\textbf{Harmonization Quality}} & \multicolumn{2}{c}{\textbf{Temporal Consistency}} & \multicolumn{3}{c}{\textbf{User Study (\%)}} \\
    \cmidrule(lr){2-5} \cmidrule(lr){6-7} \cmidrule(lr){8-10}
    & \textbf{PSNR $\uparrow$} & \textbf{SSIM $\uparrow$} & \textbf{LPIPS $\downarrow$} & \textbf{RMSE $\downarrow$} & \textbf{CLIP Score $\uparrow$} & \textbf{Motion Preservation $\downarrow$} & \textbf{Temporal $\uparrow$} & \textbf{ID $\uparrow$} & \textbf{Harmonization $\uparrow$} \\
    \midrule
    IC-Light~\cite{zhang2025iclight} & 14.77 & 0.8889 & 0.0828 & 0.1881 & 0.9895 & 1.2928 & 56 (\%)& 57 (\%)& 27 (\%)\\
    Relightful Harmonization~\cite{ren2024relightful} & 15.89 & 0.9301 & 0.0581 & 0.1643 & 0.9907 & 1.0021 & 36 (\%)& 50 (\%)& 36 (\%)\\
    RelightVid~\cite{fang2025relightvid} & 15.70 & 0.9214 & 0.0707 & 0.1711 & 0.9946 & 0.7096 & 35 (\%)& 51 (\%)& 27 (\%)\\
    Light-A-Video~\cite{zhou2025lightavid} & 15.64 & 0.8900 & 0.0791 & 0.1716 & 0.9955 & 0.5775 & 56 (\%)& 58 (\%)& 51 (\%)\\
    \textbf{Ours (HarmoVid)} & \textbf{17.91} & \textbf{0.9306} & \textbf{0.0554} & \textbf{0.1325} & \textbf{0.9963} & \textbf{0.5264} & \textbf{82 (\%)} & \textbf{78 (\%)} & \textbf{72 (\%)} \\
    \bottomrule
  \vspace{-2em}
    
  \end{tabular}%
  }
\end{table*}

\section{Experiments}
\label{sec:experiments}

We conduct comprehensive experiments to evaluate the effectiveness of our \textbf{HarmoVid} in terms of both spatial harmonization quality and temporal stability.  
Our experiments include quantitative comparisons with implementation details, existing baselines, temporal smoothing analysis, ablation studies, and user studies.  

\subsection{Experimental Details}
\label{sec:experimental_details}

\paragraph{Implementation Details.} 
We curated videos from stock video sources and applied Grounded-SAM-2~\cite{ren2024grounded} to both identify portrait content and extract masks, resulting in a total of 10,000 portrait videos. For our experiments, we sampled 100 frames per video, and reserved a subset of 200 videos for testing.

We train HarmoVid with a standard L2 diffusion loss on 8 NVIDIA A100 GPUs for 8 hours, totaling 1,200 iterations.
During inference, we employ temporal MultiDiffusion~\cite{Zhang_2024_CVPR} for videos longer than 85 frames, enabling high-quality video harmonization on extended sequences.

\paragraph{Baselines.}  
We compare our approach against recent state-of-the-art harmonization methods, including IC-Light~\cite{zhang2025iclight}, Relightful Harmonization~\cite{ren2024relightful}, RelightVid~\cite{fang2025relightvid}, and Light-A-Video~\cite{zhou2025lightavid}.
For a fair comparison, all methods are evaluated on the same composited foreground–background pairs.
Among these, IC-Light and Relightful Harmonization perform frame-wise harmonization, whereas RelightVid and Light-A-Video incorporate temporal consistency modeling.
As the original Light-A-Video does not support background conditioning, we incorporated its training-free module into the IC-Light background conditioning version.
Additionally, to identify the effectiveness of our Video Deflickering Network, we compare our lighting deflickering model with a general deflickering baseline, BVD~\cite{lei2023deflicker}.

\paragraph{Evaluation datasets.}  
We evaluate our model both quantitatively and qualitatively, focusing on spatial harmonization quality and temporal consistency.

For qualitative evaluation, we utilize a real portrait dataset, where each foreground subject is composited with various arbitrary background scenes to demonstrate the visual harmonization capability of our method.  

For quantitative evaluation, we construct two synthetic test sets derived from the same real portrait dataset by applying a color look-up-table (LUT)~\cite{lu2022lut} to the foreground videos.  
In the first setting, a \textit{single LUT is consistently applied to all frames within each video}, producing uniformly relit sequences.  
This setup provides paired videos that enable direct numerical comparison with the original unmodified sequences, allowing us to evaluate the video harmonization quality (\Cref{sec:harmonization_comparison}).  
In the second setting, \textit{different LUTs are randomly applied on a per-frame basis} to introduce temporal color and illumination variations, simulating flickering artifacts.  
This dataset is used to evaluate the effectiveness of our Video Deflickering Network in restoring temporal consistency while maintaining appearance fidelity (\Cref{sec:deflicker_comparison}). 
Together, these two complementary datasets allow a comprehensive evaluation of both the harmonization and deflickering performance of our proposed method.

\paragraph{Evaluation metrics.}
\textit{Temporal consistency} is measured using two complementary metrics used in Light-A-Video~\cite{zhou2025lightavid}. First, we calculate the average CLIP~\cite{radford2021clip} similarity between consecutive frames to capture semantic stability over time (CLIP Score). Second, we estimate optical flow with RAFT~\cite{teed2020raft} and compute the flow discrepancy between the harmonized and reference videos, reflecting how well motion dynamics are preserved (Motion Preservation). 

For \textit{frame-wise harmonization quality}, we use PSNR, SSIM, LPIPS, and RMSE by comparing the generated frames with corresponding reference videos. Higher PSNR and SSIM, along with lower LPIPS and RMSE, indicate better perceptual fidelity and realism, while higher CLIP Score and lower Motion Preservation indicate better spatial and temporal consistency.

\begin{figure*}[ht]
\centering
\scriptsize

\vspace{1mm}
\begin{tikzpicture}
    \draw[->, thick] (0,0) -- (8.2,0); 
    \draw[->, thick] (8.3,0) -- (15.5,0) ;
    \node[anchor=west] at (15.7,0) {\textbf{Time}};
\end{tikzpicture}

\newcommand{\imagerow}[5]{%
  \begin{minipage}[c]{0.01\textwidth}
    \centering
    \raisebox{-0.5em}{\makebox[0pt][r]{\rotatebox{90}{\textbf{#1}}}}
  \end{minipage}%
  \begin{minipage}[c]{0.98\textwidth}
    \centering
    \includegraphics[width=0.24\linewidth,height=0.1371\linewidth]{#2}%
    \hspace{0.2mm}%
    \includegraphics[width=0.24\linewidth,height=0.1371\linewidth]{#3}%
    \hspace{0.6mm}%
    \includegraphics[width=0.24\linewidth,height=0.1371\linewidth]{#4}%
    \hspace{0.2mm}%
    \includegraphics[width=0.24\linewidth,height=0.1371\linewidth]{#5}%
  \end{minipage}\par\vspace{0.1mm}
}

\imagerow{Input}
  {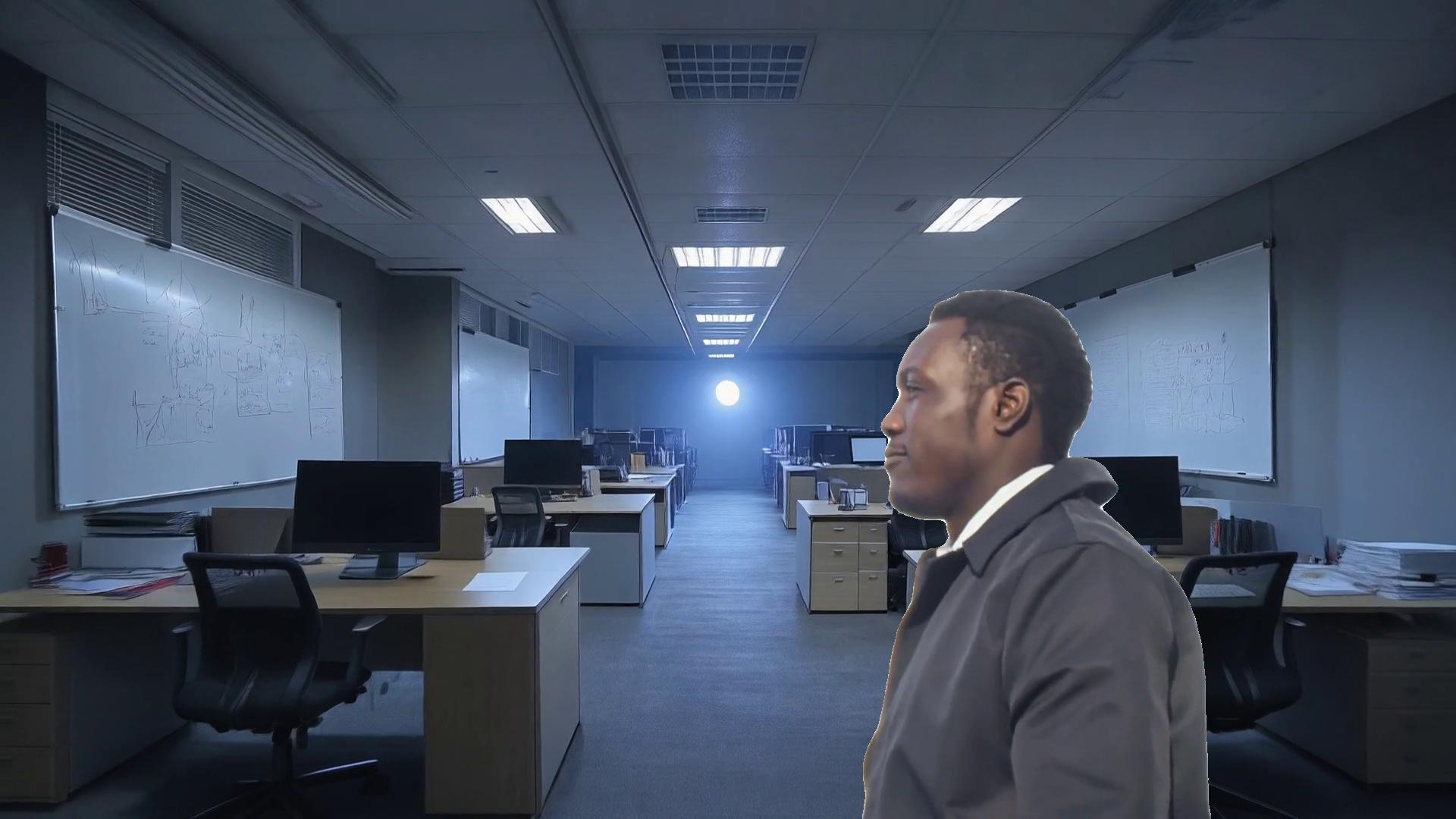}
  {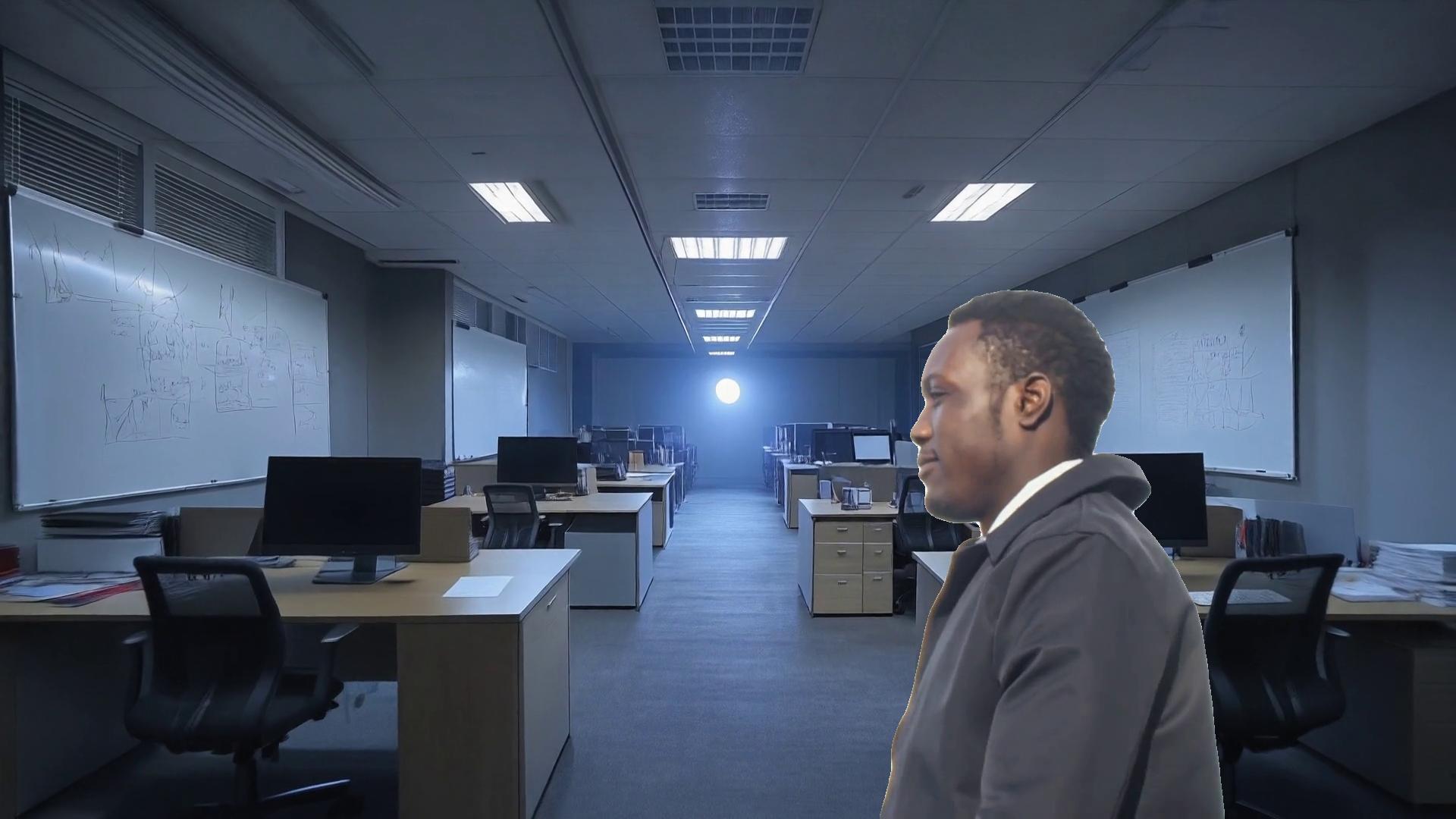}
  {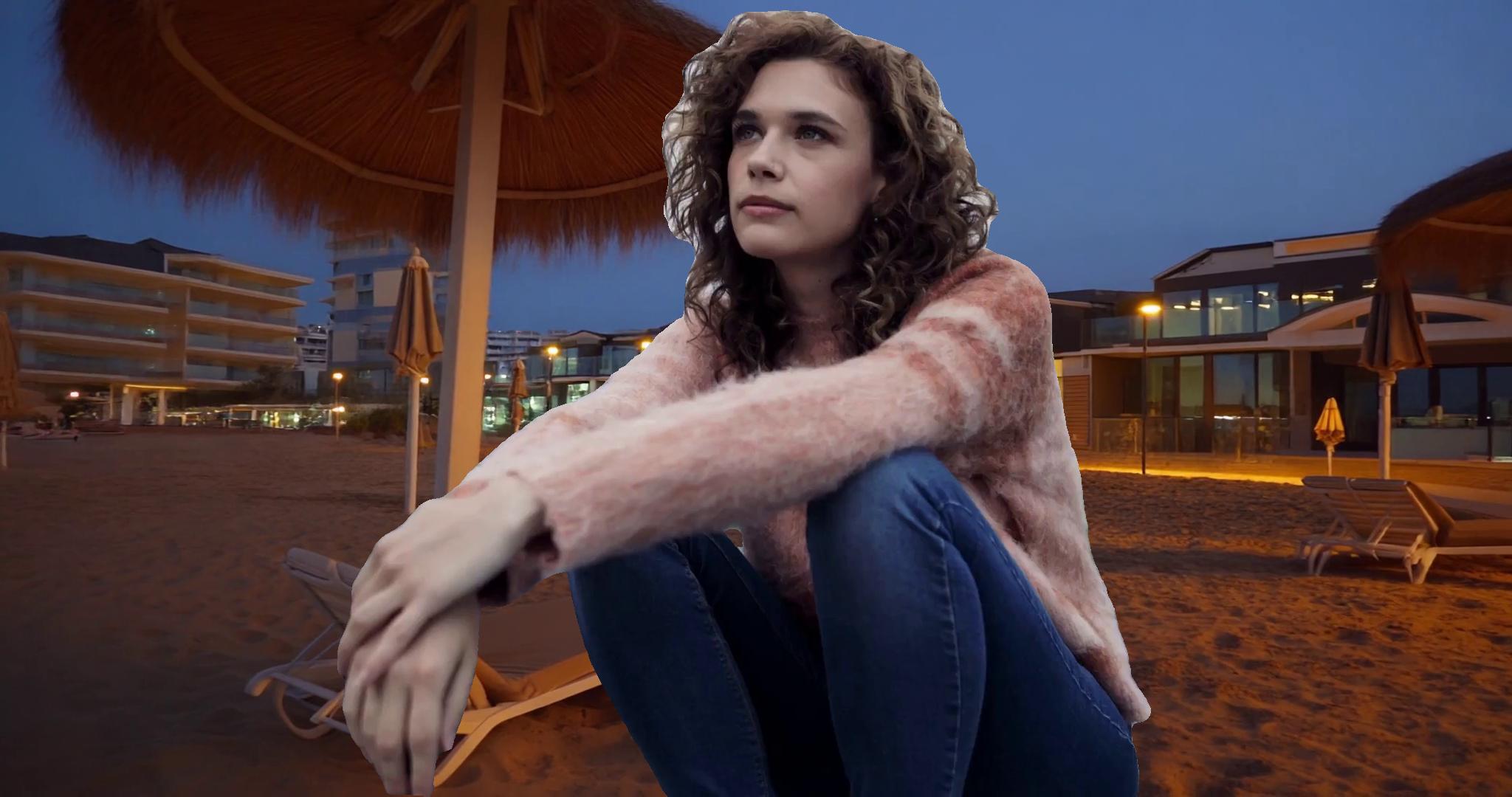}
  {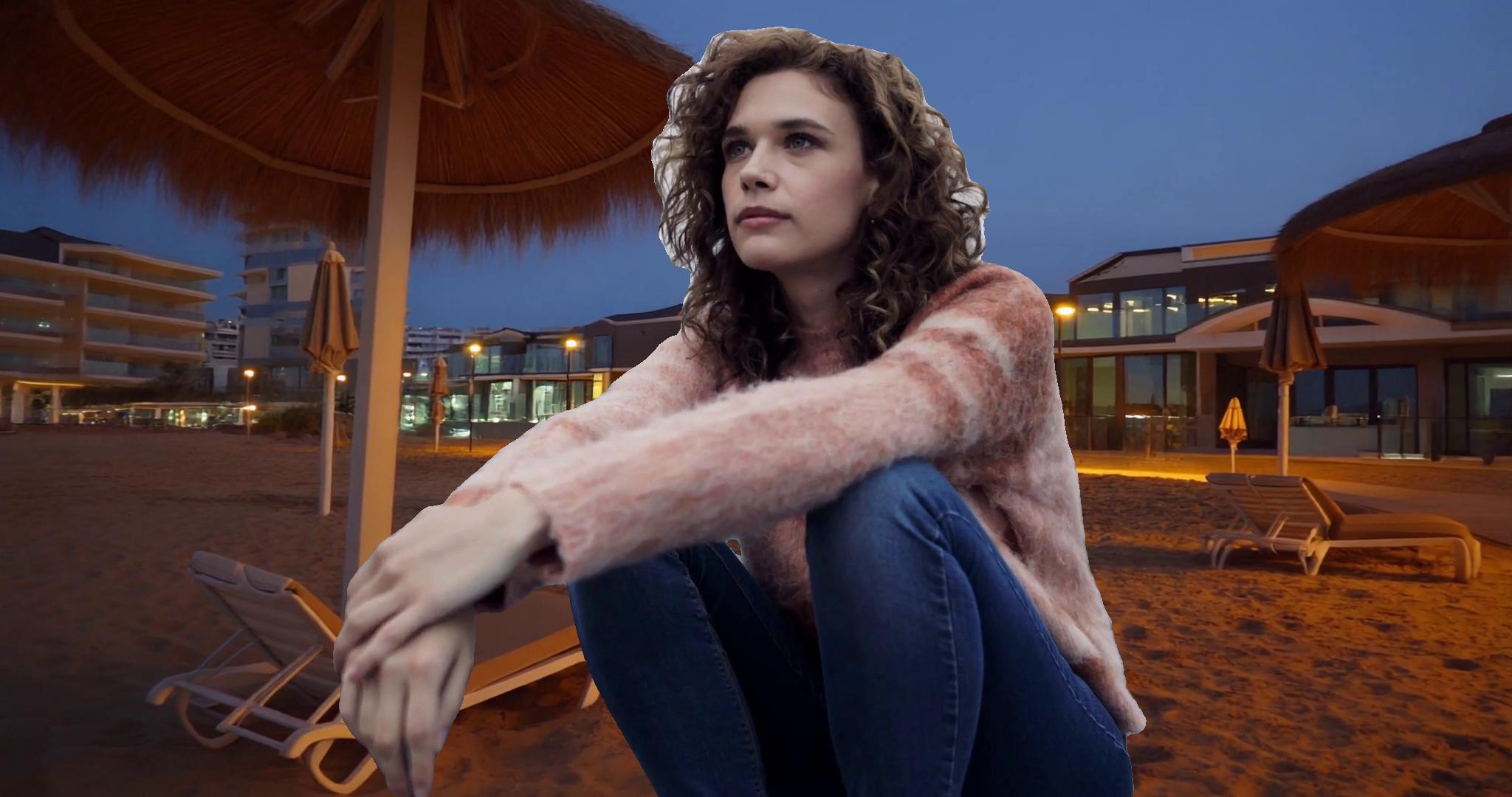}

\imagerow{IC-Light}
  {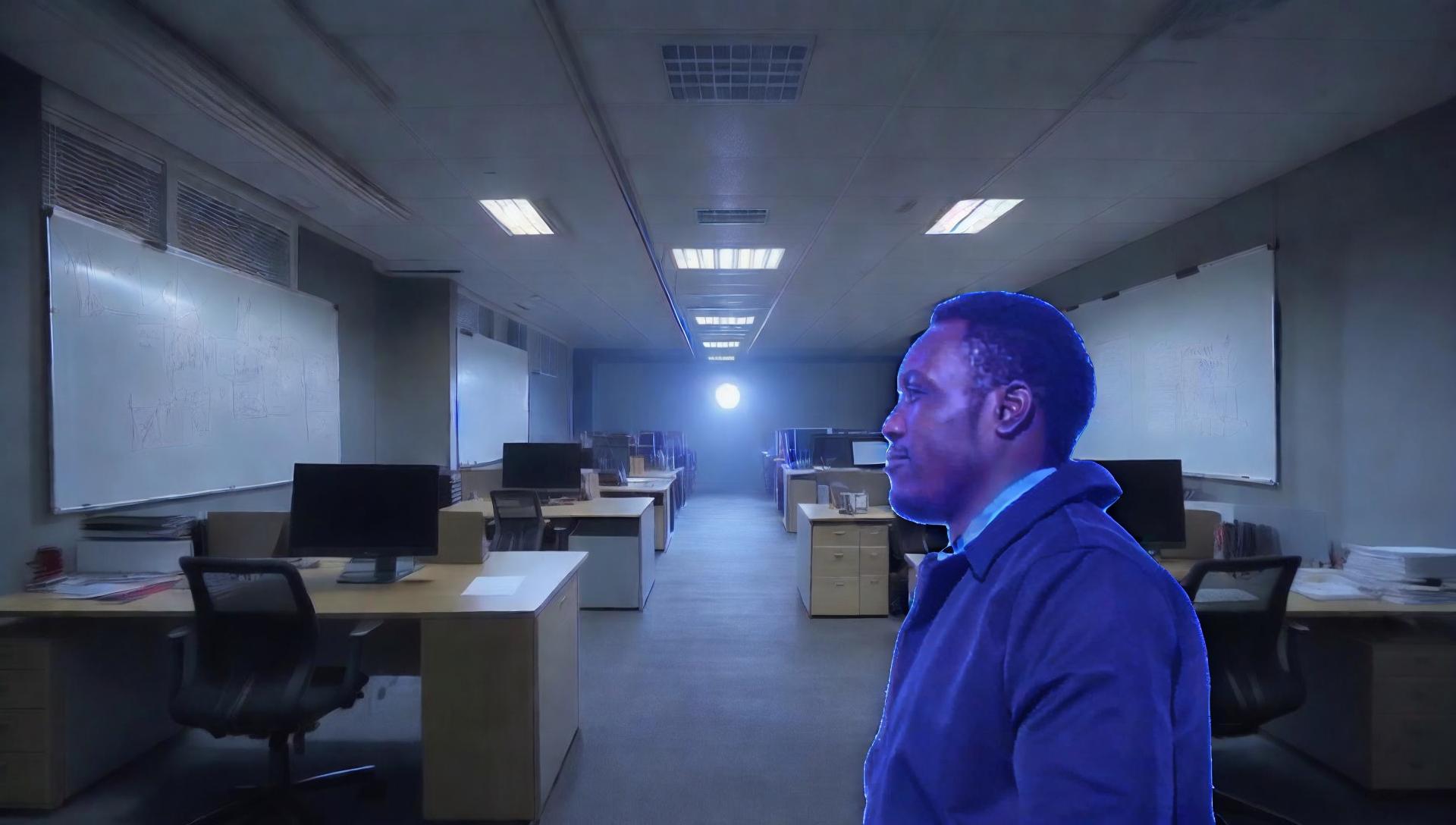}
  {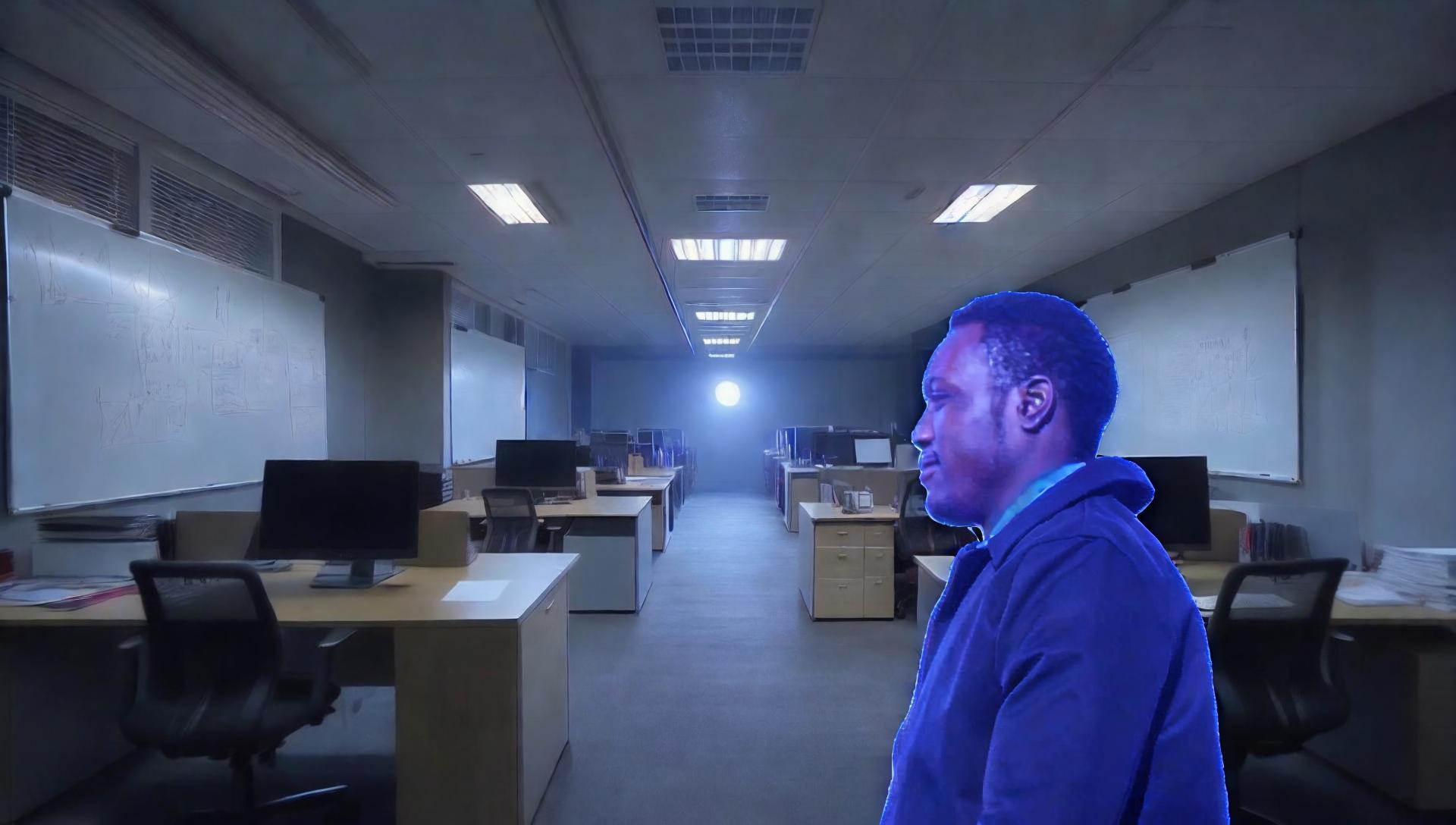}
  {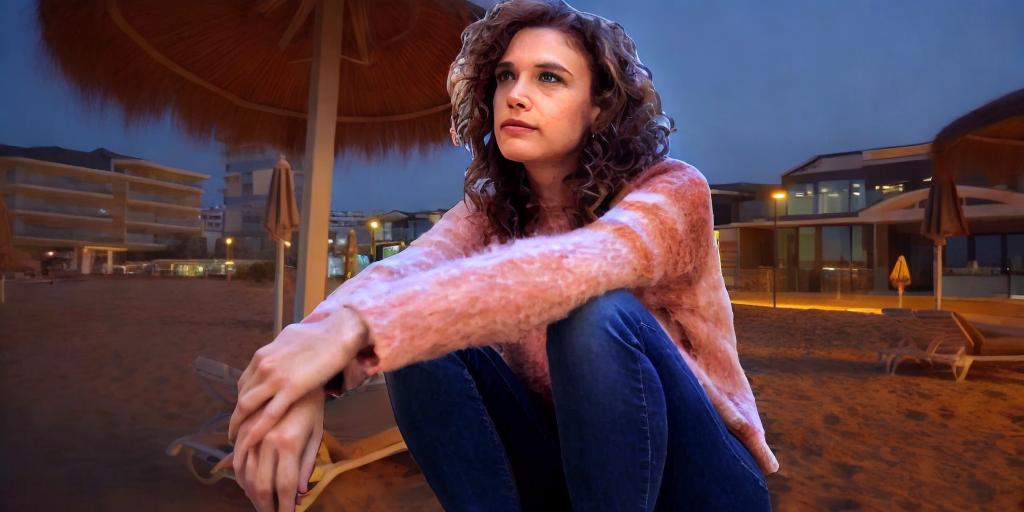}
  {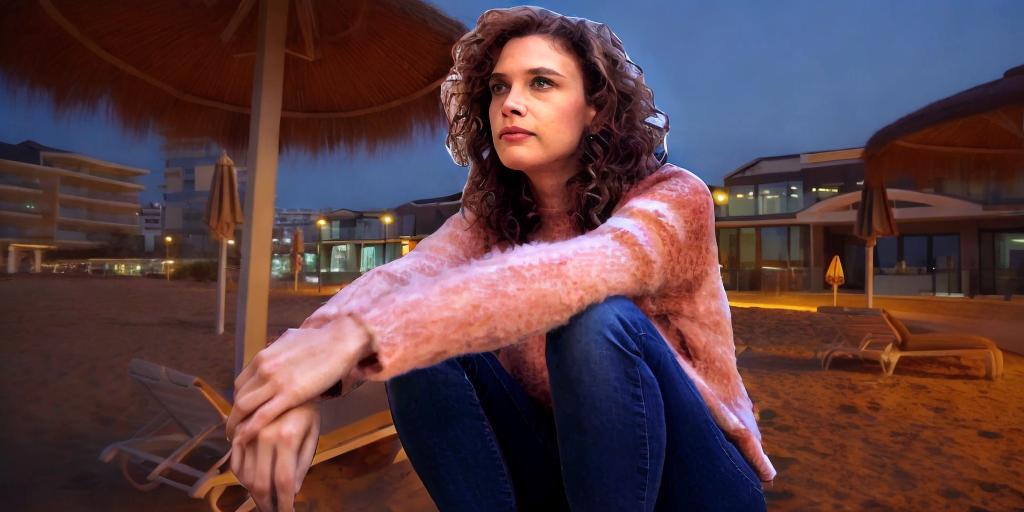}

\imagerow{Rel Harm}
  {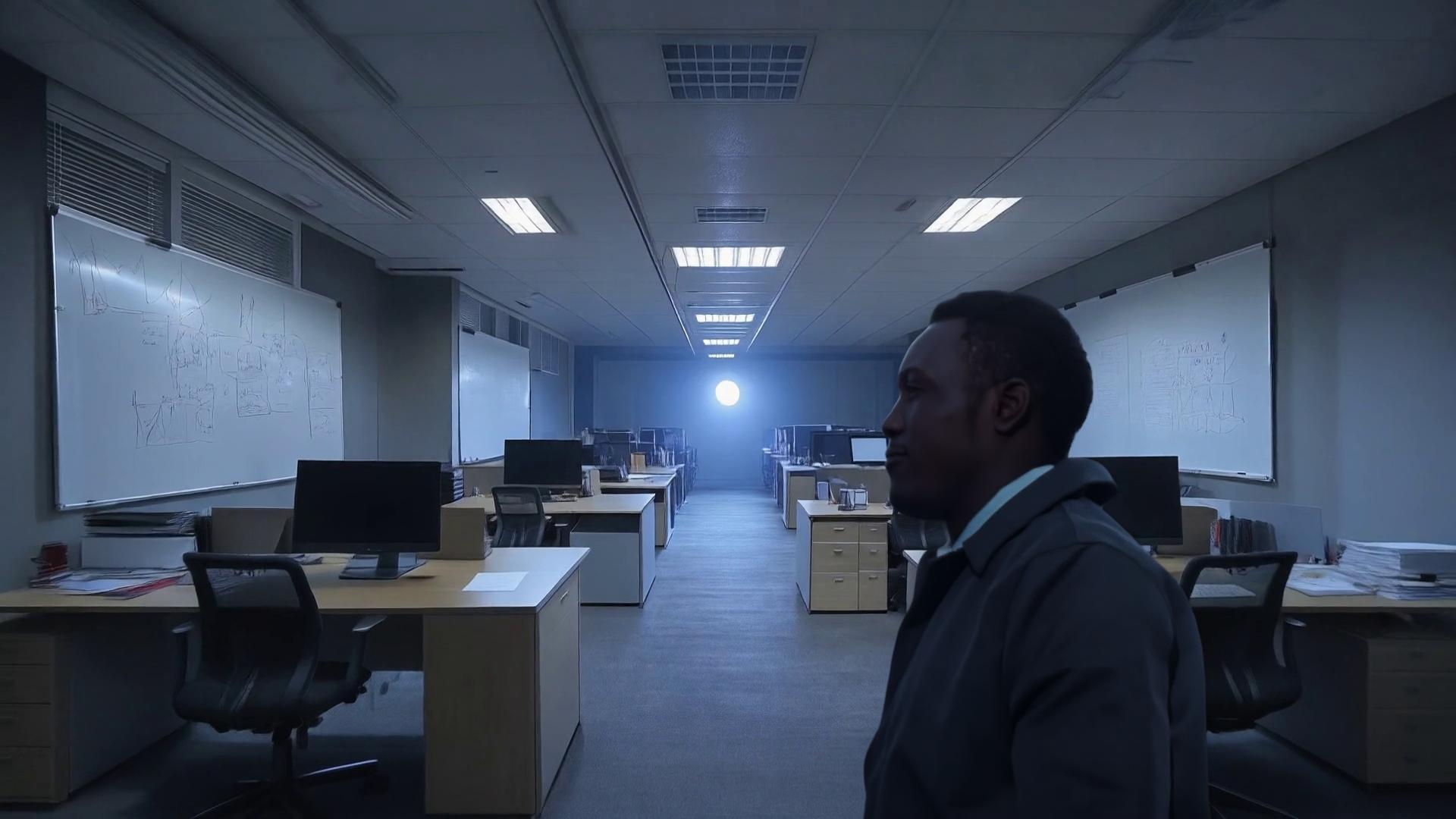}
  {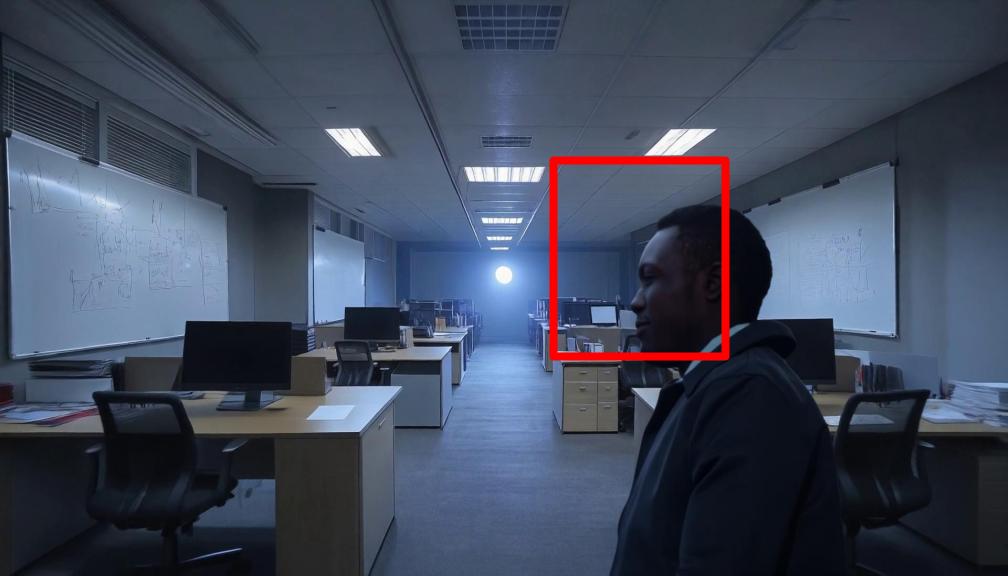}
  {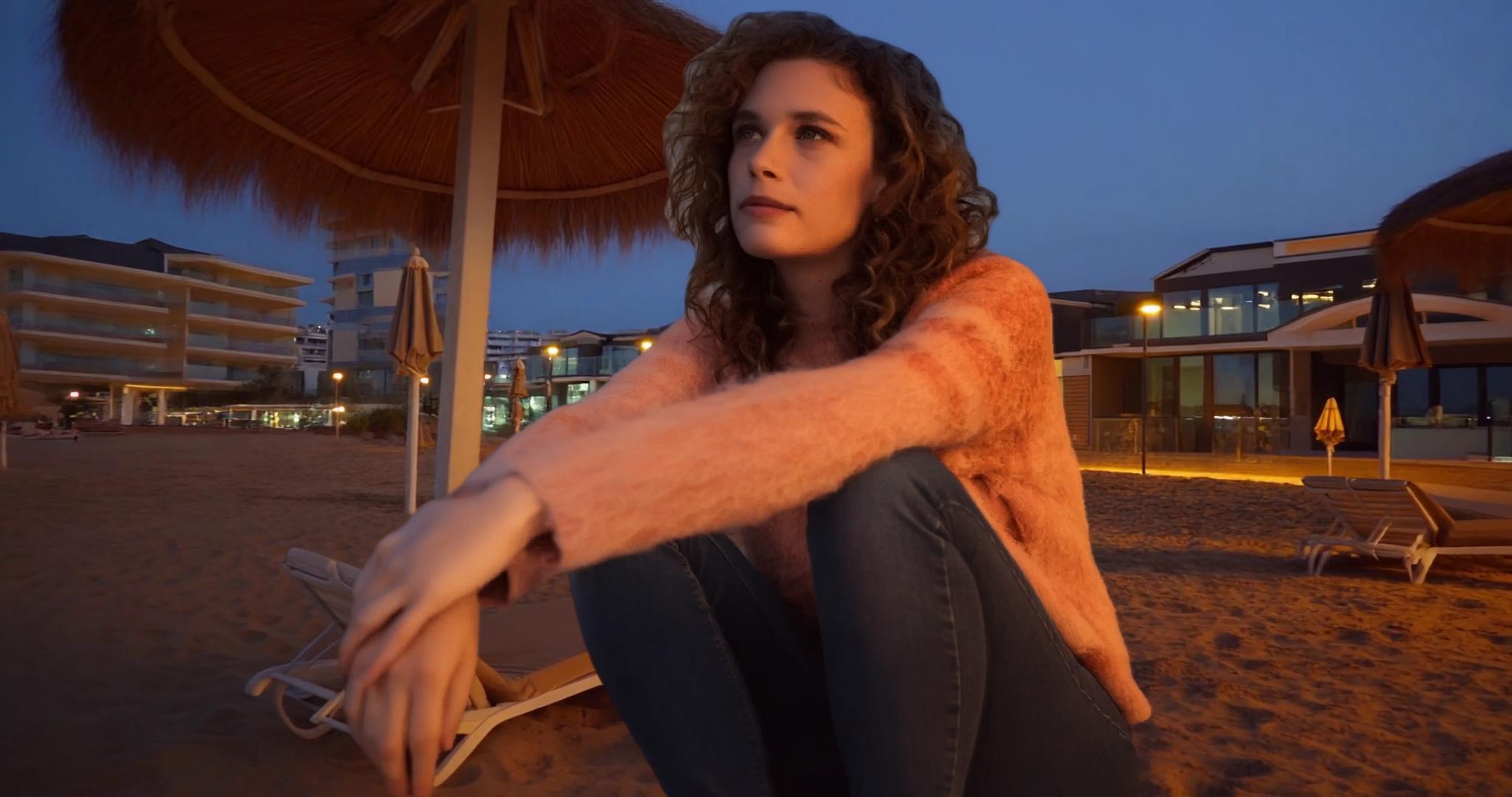}
  {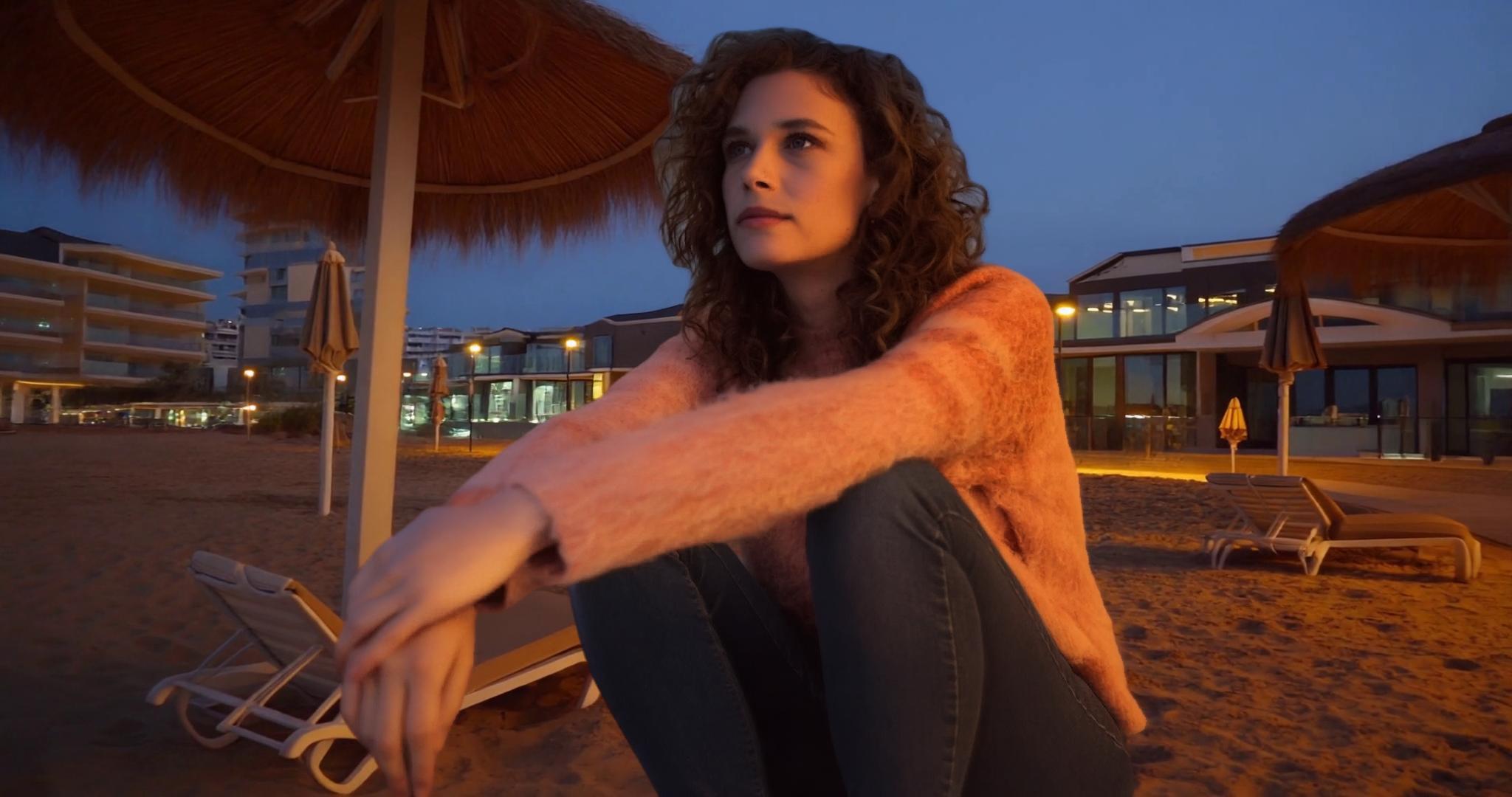}

\imagerow{RelightVid}
  {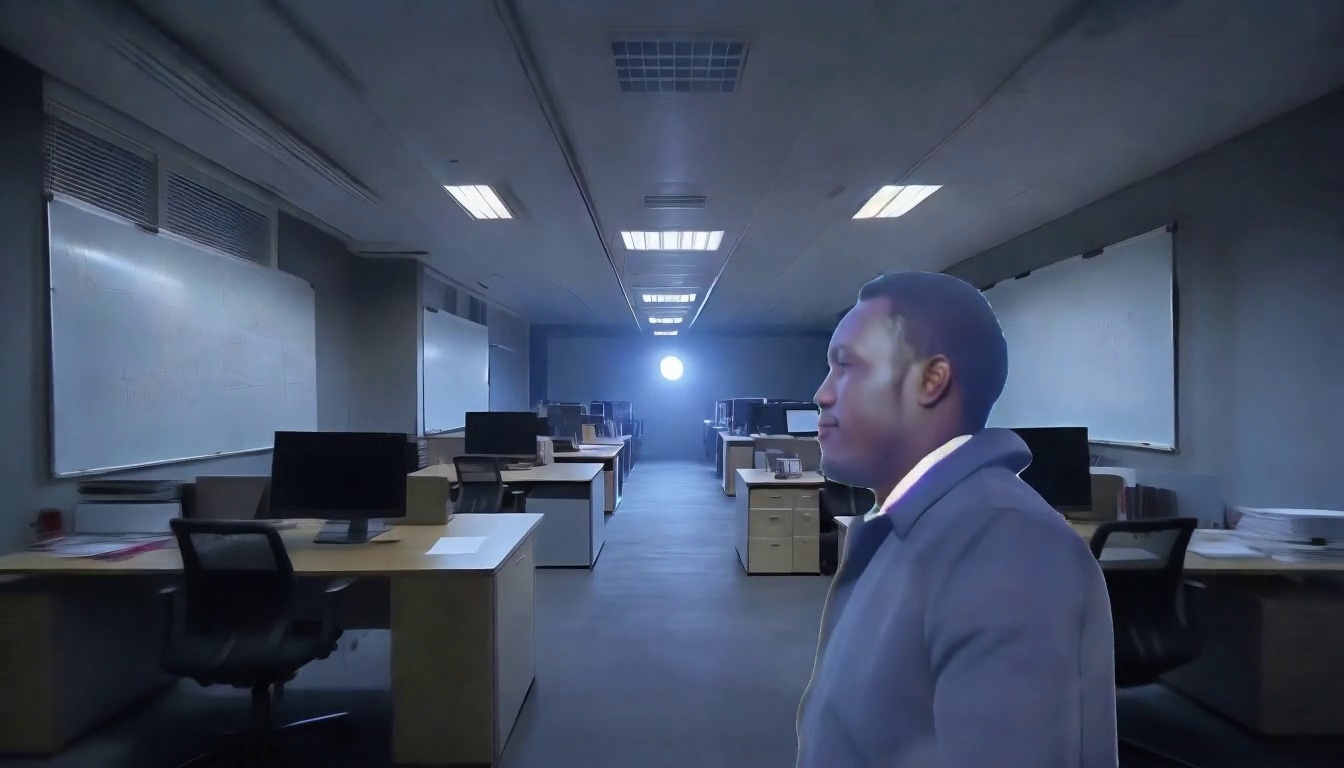}
  {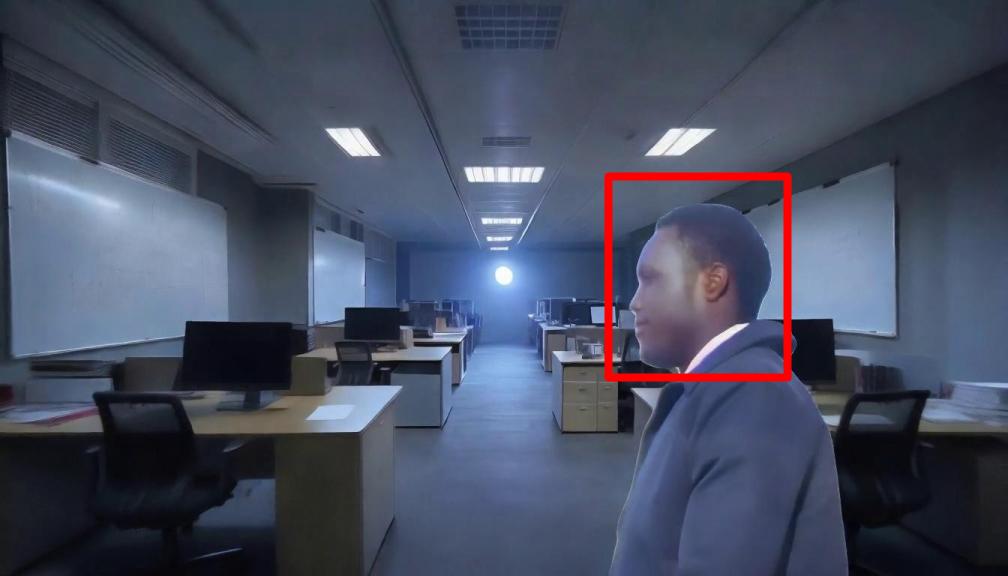}
  {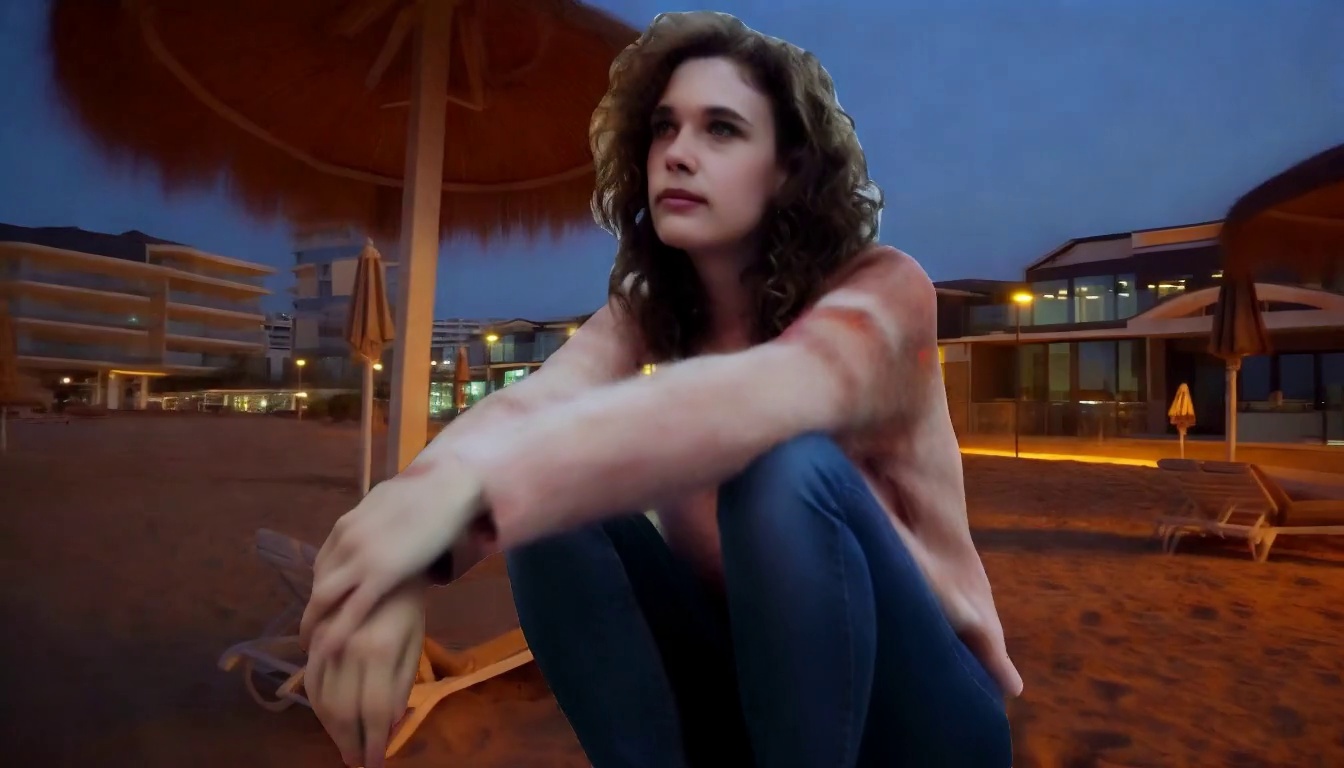}
  {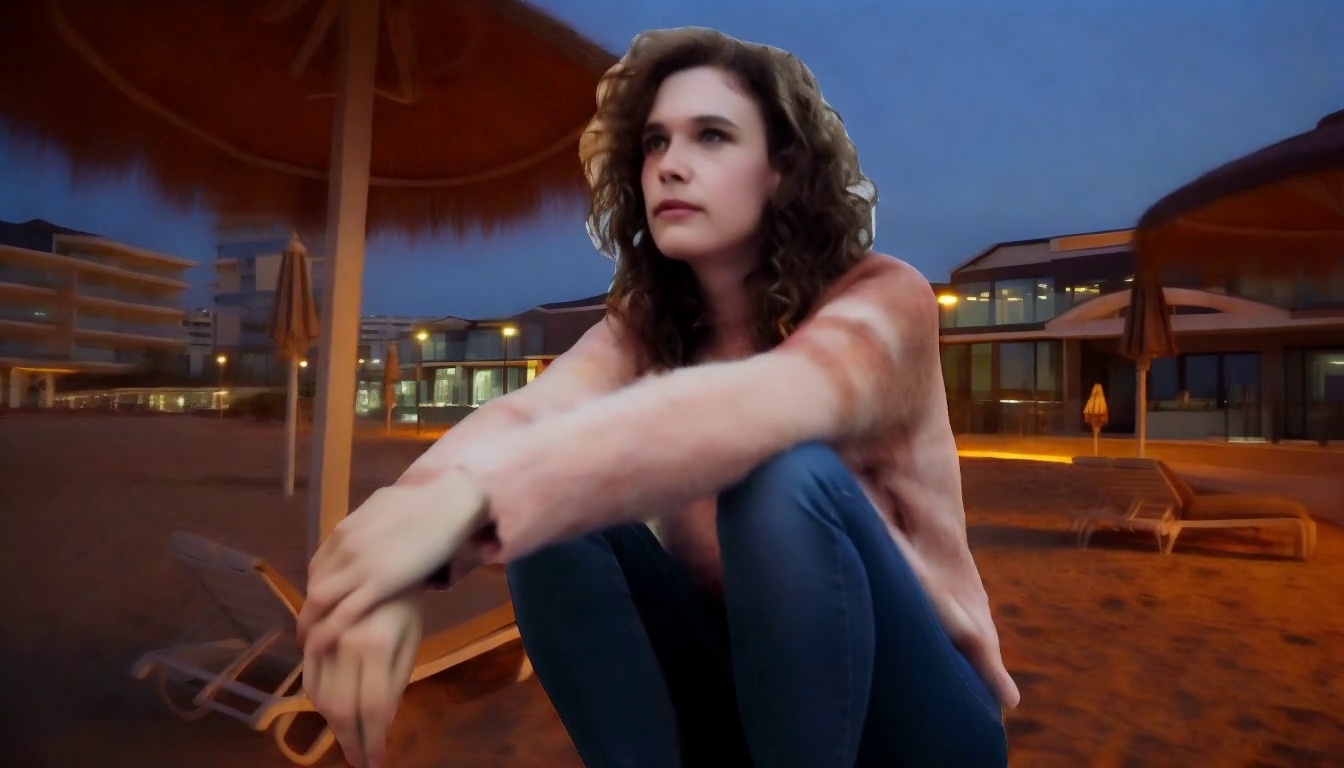}

\imagerow{Light-A-Video}
  {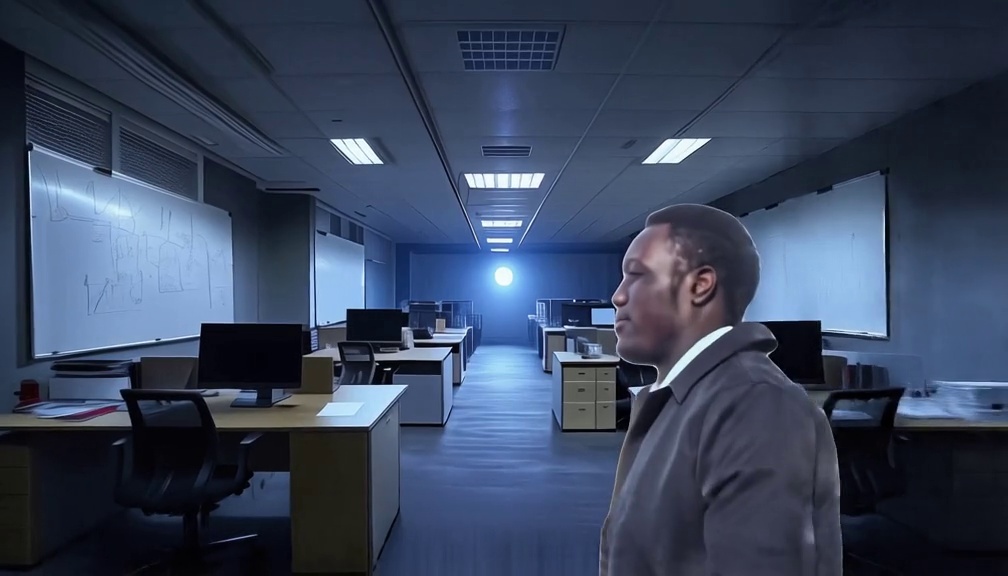}
  {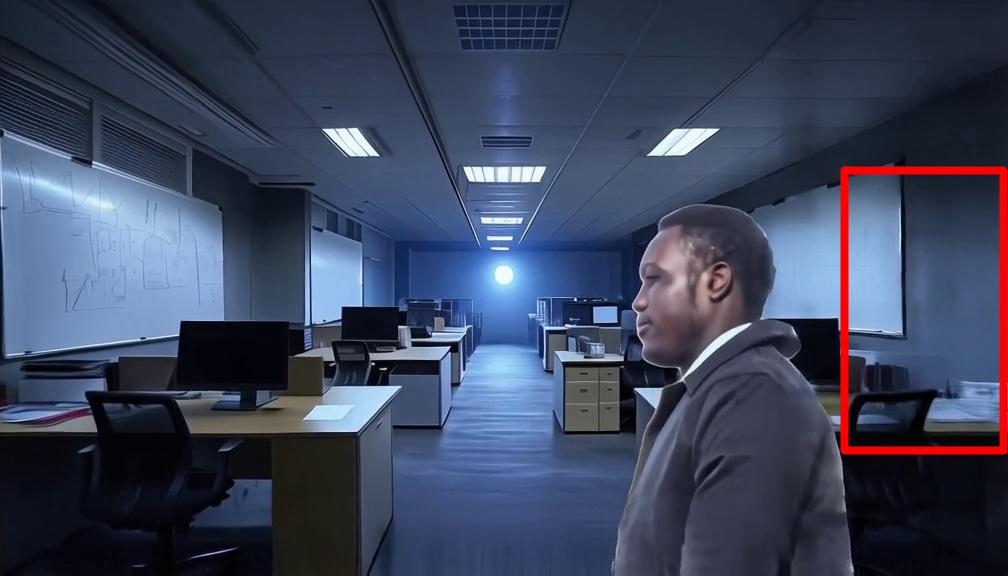}
  {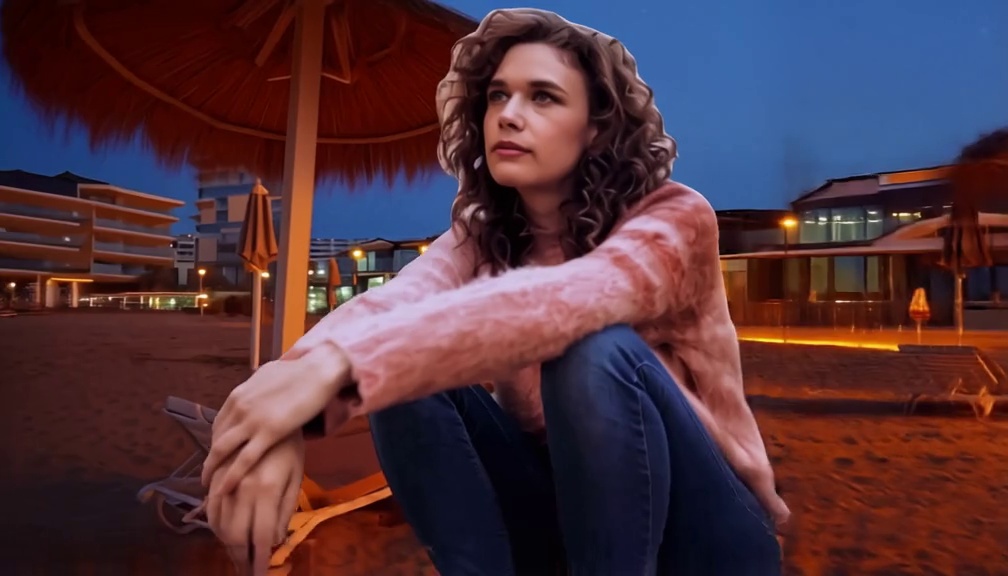}
  {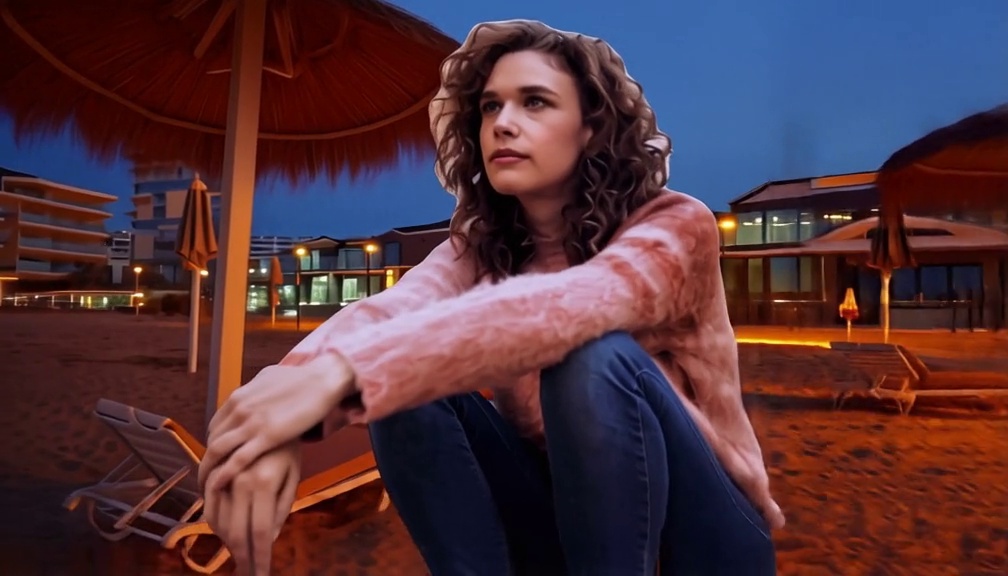}

\imagerow{Ours}
  {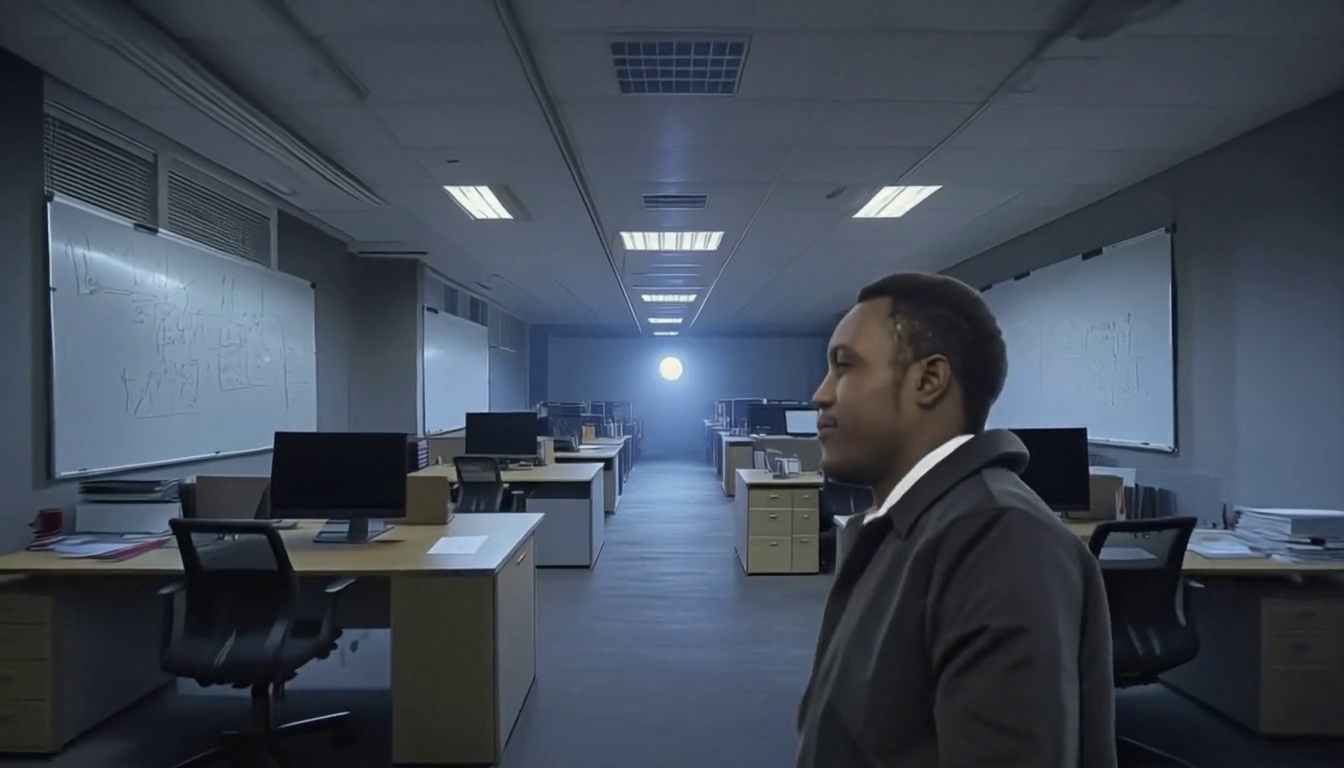}
  {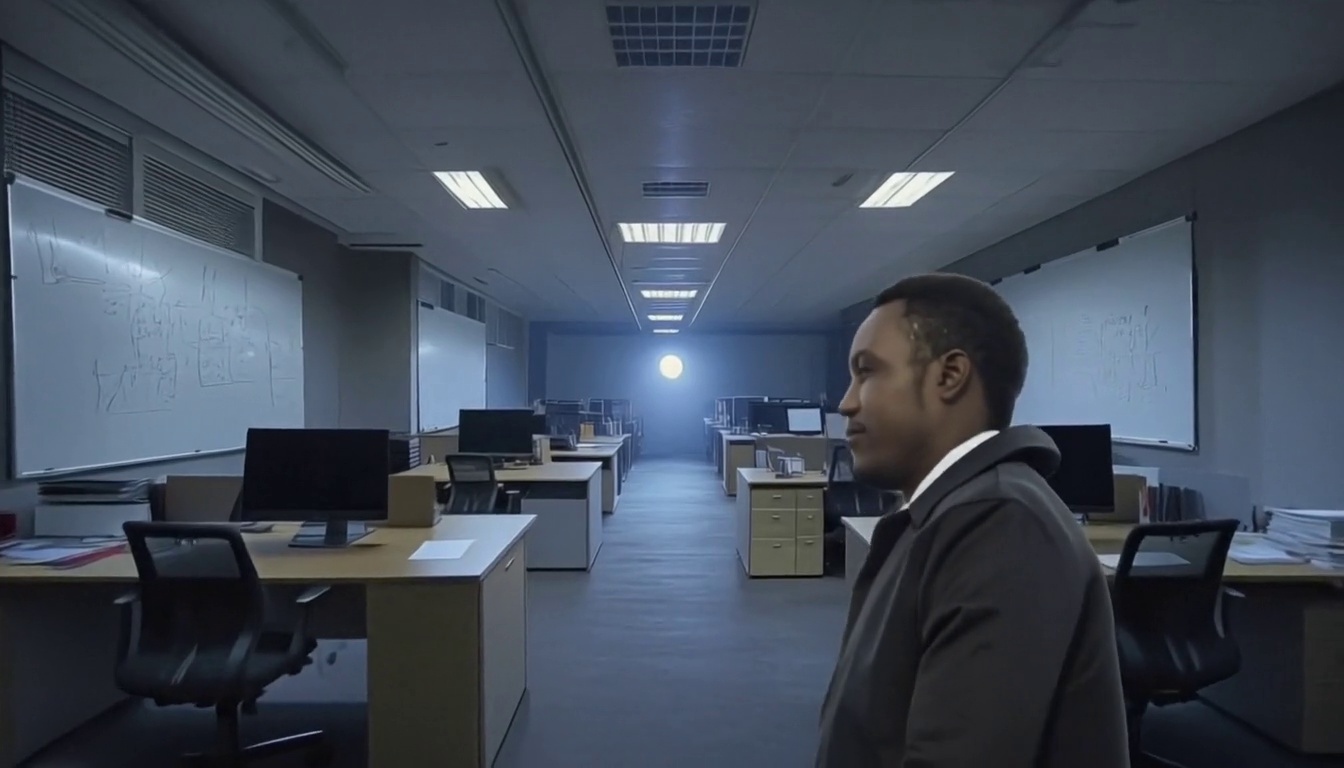}
  {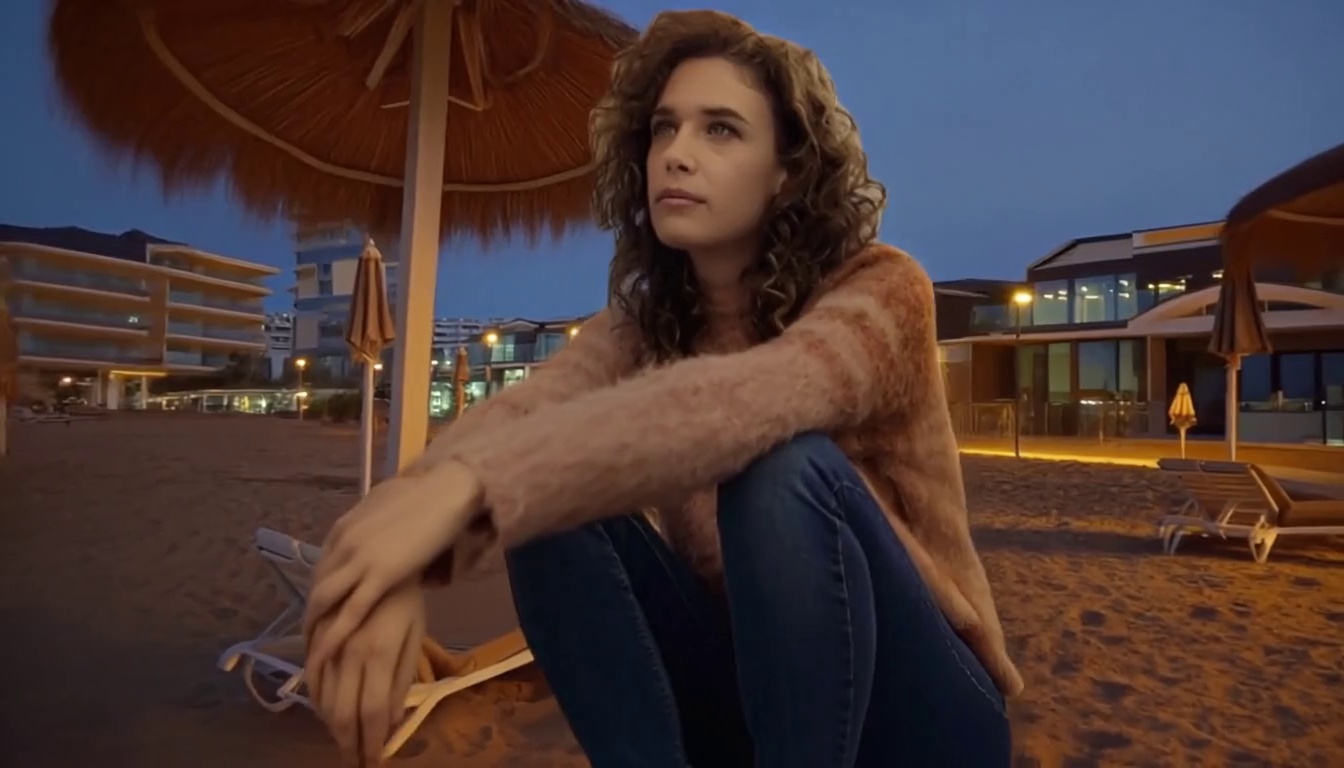}
  {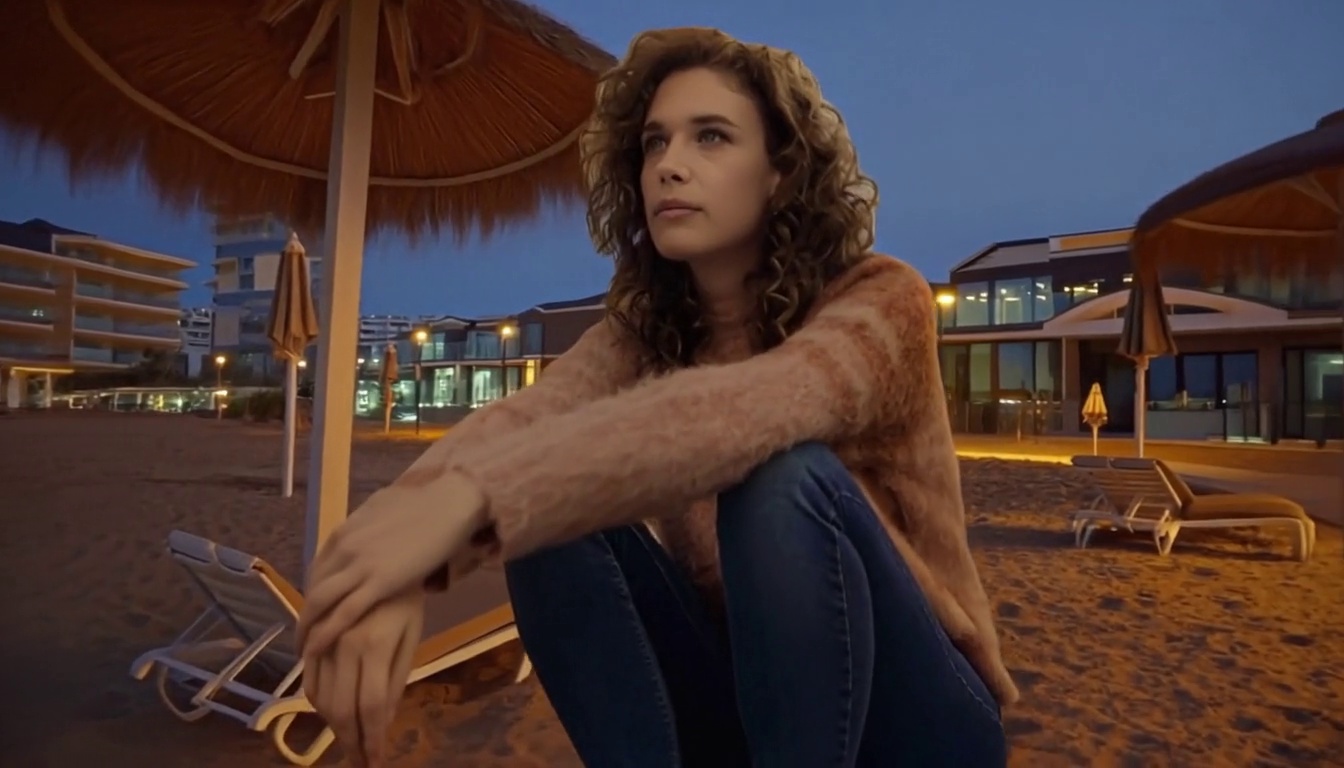}

\vspace{-1mm}
\caption{
Qualitative comparison of our HarmoVid to other image- or video-based video harmonization methods.
}
\label{fig:main_figure}
\vspace{-1.5em}
\end{figure*}

\subsection{Video Harmonization Comparison}
\label{sec:harmonization_comparison}
We comprehensively evaluate the proposed \textbf{HarmoVid} framework against state-of-the-art video harmonization methods. As shown in \Cref{tab:main_comparison}, HarmoVid consistently achieves superior spatial and temporal harmonization performance compared to existing approaches. Qualitative comparisons in \Cref{fig:main_figure} further demonstrate that our method seamlessly blends the foreground and background while maintaining realistic illumination and temporal stability. In particular, in the left examples, IC-Light tends to alter global colors excessively, Relightful Harmonization exhibits reflection flickering, RelightVid struggles with ID preservation, and Light-A-Video introduces artifacts in the background. In contrast, our results remain temporally consistent and visually realistic. The right example also highlights that other methods often produce noticeable artifacts around the portrait region, whereas our approach effectively suppresses them, thanks to the use of a pseudo alpha mask that enables more precise boundary blending between the foreground and background.

\paragraph{User Study.}
We further validate the perceptual quality through a user study involving 33 participants. Each participant views 10 video clips randomly selected from a total of 30 samples. For each clip, participants evaluate results from five competing methods by selecting \textit{all} methods they consider \textit{appropriate} in terms of:
(1) \textit{Temporal Consistency} — smoothness and stability of motion,
(2) \textit{ID Preservation} — consistency of subject identity and appearance, and
(3) \textit{Overall Harmonization} — naturalness and lighting realism.
The selections are then aggregated across participants to derive subjective preference scores, complementing the quantitative evaluation metrics.

The aggregated responses in \Cref{tab:main_comparison} show that HarmoVid is the most frequently preferred, confirming its advantage in producing visually coherent and temporally stable results.

\begin{table}[t]
  \centering
  \includegraphics[width=\linewidth]{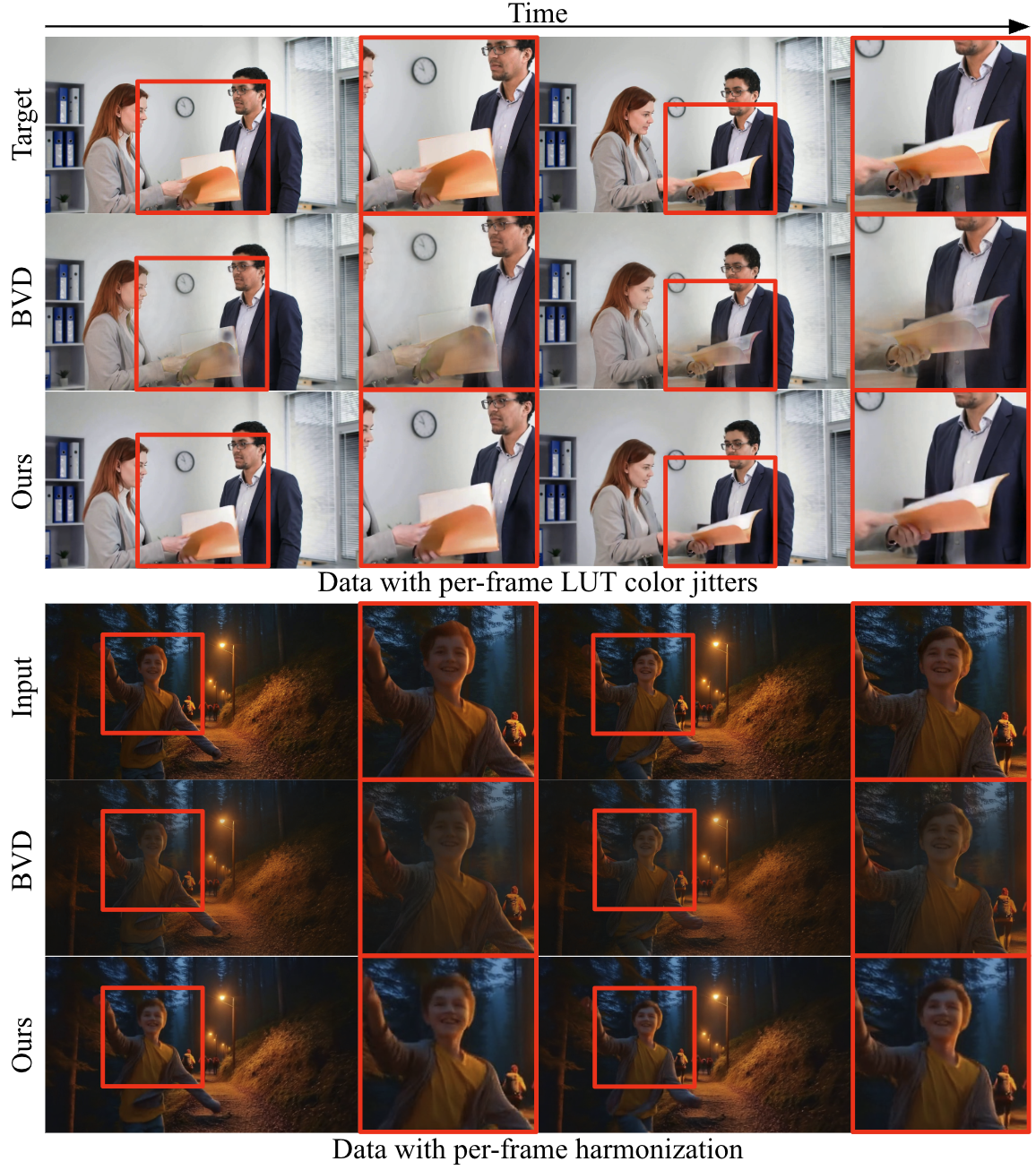} \\
  \resizebox{\linewidth}{!}{%
  \begin{tabular}{lcccc}
    \toprule
    & \multicolumn{2}{c}{\bfseries Per-frame LUT color jitters} & \multicolumn{2}{c}{\bfseries Per-frame harmonization}\\
    \cmidrule(lr){2-3} \cmidrule(lr){4-5}
    & {\bfseries CLIP Score $\uparrow$} & {\bfseries Motion Pres. $\downarrow$} & {\bfseries CLIP Score $\uparrow$} & {\bfseries Motion Pres. $\downarrow$}\\
    \midrule
    BVD~\cite{lei2023deflicker} 
    & 0.9950 & 0.5114 & 0.9920 & 1.3439   \\
    HarmoVid  
    & \textbf{0.9967} & \textbf{0.3630} & \textbf{0.9936} & \textbf{0.5395}   \\
  \bottomrule
\end{tabular}%
}
\caption{We demonstrate the effectiveness of our deflickering network.
In the upper synthetic example, BVD suffers from noticeable loss of spatial detail, whereas in the lower per-frame harmonization example, it fails to preserve lighting effects. Our lighting deflickering method effectively stabilize the global and local lighting flickering artifacts with minimum loss of details and lightings.
}
\vspace{-1em}

\label{tab:deflicker}
\end{table}

\subsection{Deflickering Comparison}
\label{sec:deflicker_comparison}

We evaluate the effectiveness of the proposed Video Deflickering Network in reducing temporal artifacts in flickering videos. Specifically, we compare our deflickering results with BVD~\cite{lei2023deflicker} on the video data applied with per-frame LUT color jitters described in \Cref{sec:experimental_details}, which mainly exhibit typical frame-to-frame flickering. We further apply our deflickering method to the video data applied with per-frame harmonization outputs that exhibit more complex temporal inconsistencies of local lighting, such as cast-shadow and highlight flickering.
For quantitative evaluation, we employ the same motion preservation metric; however, due to the absence of ground truth videos, the metric is computed using the foreground regions of the original frames. Evaluations based on CLIP Score and Motion Preservation indicate that our method improves temporal consistency in both settings (\Cref{tab:deflicker}). Moreover, qualitative comparisons on consecutive frames demonstrate that our method effectively suppresses flickering artifacts while preserving spatial details, including lighting effects, thereby improving temporal coherence without compromising visual fidelity.

\begin{table}
    \centering
    \includegraphics[width=\linewidth]{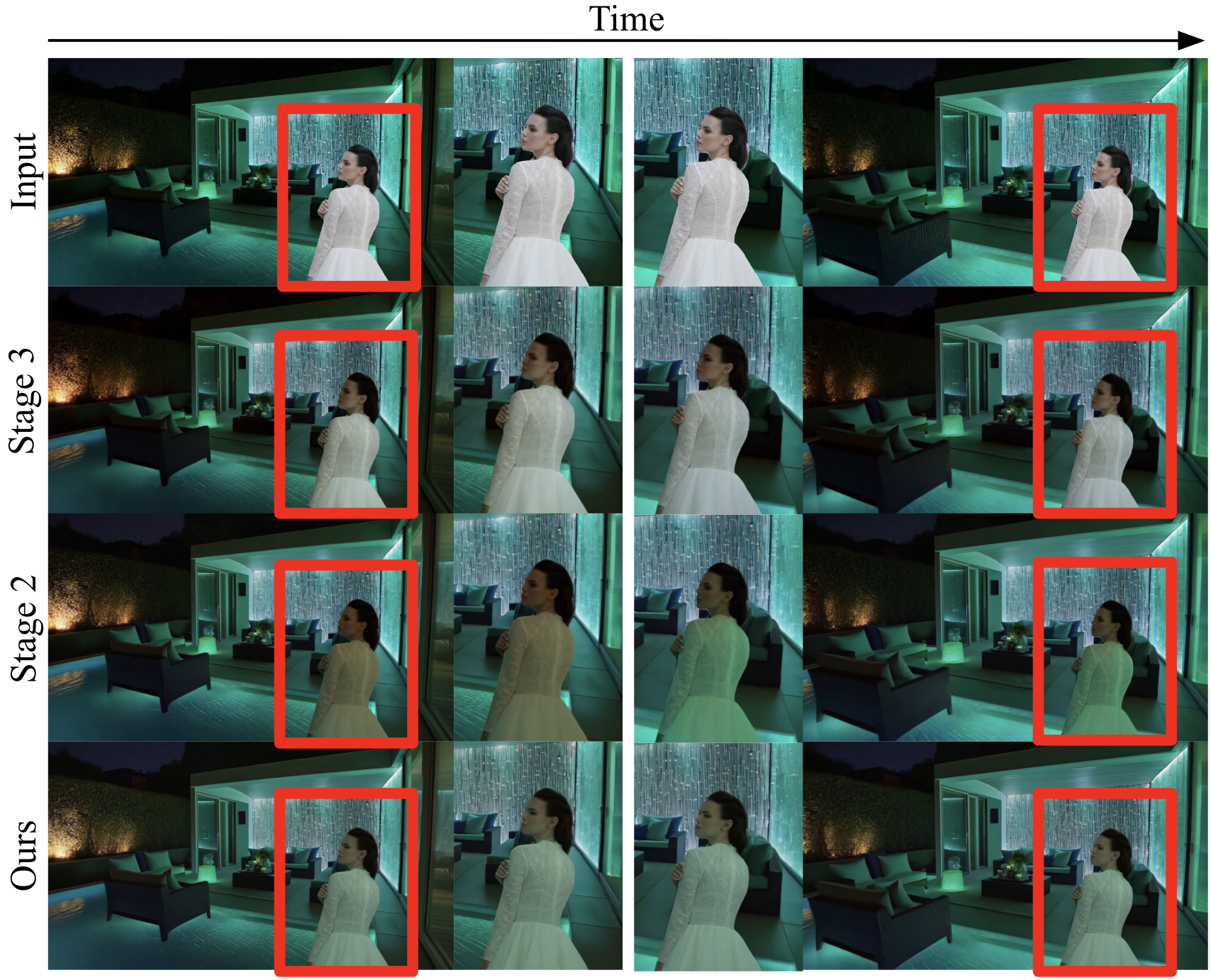} \\
    \resizebox{1\linewidth}{!}{%
    \begin{tabular}{cccccc}
        \toprule
         Stage 2 & Stage 3 & \bfseries{SSIM $\uparrow$} & \bfseries{LPIPS $\downarrow$} & \bfseries{CLIP Score $\uparrow$} & \bfseries{Motion Preservation $\downarrow$} \\
        \midrule
        \textendash & \checkmark & 0.9187 & 0.0613 & 0.9911 & 0.9376 \\
        \checkmark & \textendash & 0.9217 & 0.0594 & 0.9937 & 0.5490\\
        \checkmark & \checkmark & \textbf{0.9306} & \textbf{0.0554} & \textbf{0.9963} & \textbf{0.5264} \\
        \bottomrule
    \end{tabular}
    }
\vspace{-0.5em}
    \caption{\textbf{Ablation study}: We demonstrate the necessity of both Stage 2 and Stage 3. The results in the table are obtained using the same dataset used in \ref{sec:harmonization_comparison}, while the figures show examples applied to real-world data.}
    \vspace{-1.5em}
\label{tab:ablations_stage}
\end{table}

\subsection{More Analysis}
\paragraph{Ablation study on two-stage training.}  
To investigate the contribution of each stage, we run ablations focusing on Stage 2 (lighting deflickering; \cref{sec:deflicker}) and Stage 3 (dual-path training; \cref{sec:harmonization}), with results summarized in \cref{tab:ablations_stage}. Skipping Stage 2 degrades both temporal stability and harmonization quality: when one side of a training pair flickers, a DiT that jointly models space–time struggles to learn stable temporal representations and high-quality harmonization. In particular, \cref{tab:ablations_stage} 2nd-row, shows that harmonization is almost entirely absent. Using only the Stage 2 deflickering network for harmonization (bypassing Stage 3) suppresses short-term flicker but yields long-horizon inconsistencies (\cref{tab:ablations_stage}, 3rd-row), such as gradual illumination and tone drift; absent joint learning on real videos, outputs also appear less natural. Moreover, during inference, Stage 2 cannot be used in isolation, as it requires per-frame processing via Stage 1 (image harmonization) beforehand. Overall, the deflickering process (Stage 2) and the dual-path training (Stage 3) are complementary: together they refine fine-grained spatial details and preserve both short- and long-term temporal coherence.

\paragraph{Ablation study on pseudo-alpha mask.} 
Next, we evaluate the effect of incorporating the pseudo alpha mask during training by comparing the outputs generated with and without it. Integrating the pseudo alpha mask enhances foreground-background blending, particularly along object boundaries, leading to more natural and visually coherent harmonization.

To quantitatively assess the improvements without relying on ground-truth references, we compute reference-free image quality metrics—Laplacian Variance~\cite{memon2016imagequalityassessmentperformance} and Tenengrad—focused specifically on the foreground-background boundaries. For each frame, we extract a narrow boundary region around the pseudo alpha mask using morphological erosion and collision operations, and calculate the metrics only within these boundary pixels. Laplacian Variance evaluates edge intensity, capturing how well fine details are preserved along the boundary, while Tenengrad assesses gradient magnitudes, reflecting the clarity of structural details in complex regions. Using these boundary-focused metrics, we can systematically verify that integrating the pseudo alpha mask preserves subtle edges and maintains high-quality foreground-background blending.
As illustrated in \Cref{tab:ablations_alpha}, the benefits are most pronounced in challenging regions such as hair, where intricate details and subtle boundaries are retained more faithfully.

\begin{figure}[t]
  \centering
  \includegraphics[width=\linewidth]{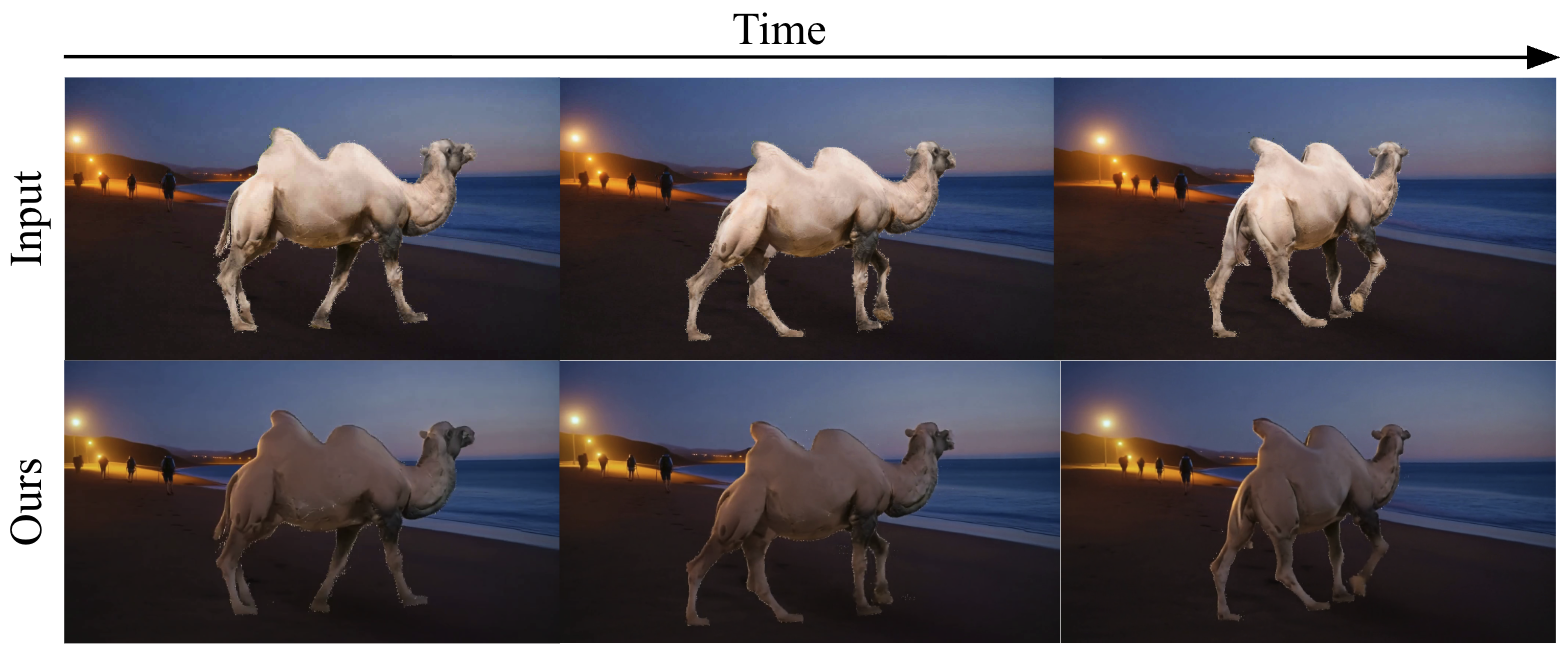}
  \caption{Result on non-portrait foreground object.}
  \vspace{-1.5em}
  \label{fig:obj}
\end{figure}

\paragraph{Generalization.} \Cref{fig:obj} shows that our model, trained exclusively on portrait foreground videos, works effectively on non-portrait foreground objects.

\begin{table}
    \centering
    \includegraphics[width=\linewidth]{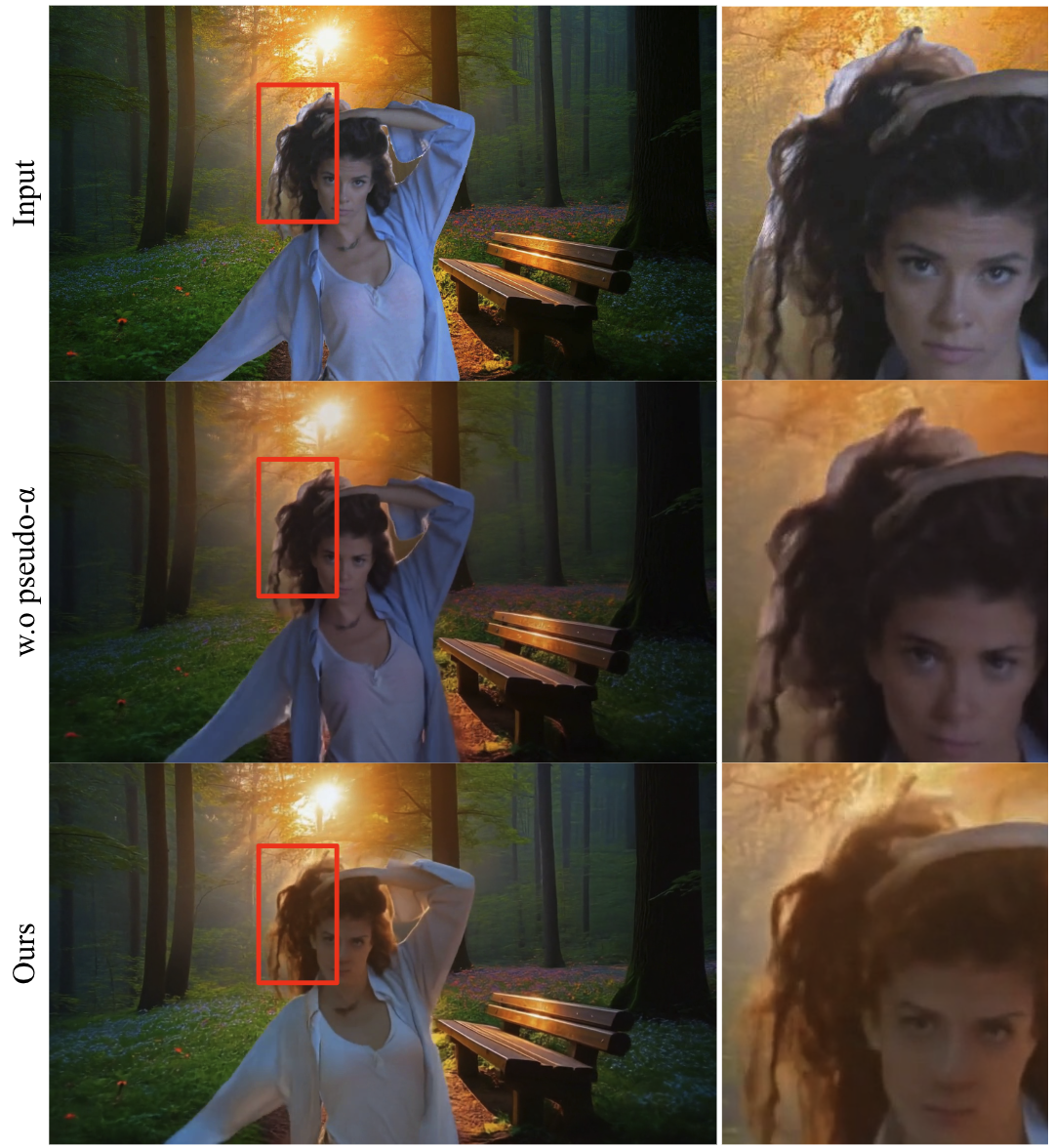} \\
    \resizebox{0.7\linewidth}{!}{%
    \begin{tabular}{ccc}
        \toprule
         pseudo-$\alpha$ & \bfseries{LaplacianVar $\uparrow$} & \bfseries{Tenengrad $\uparrow$} \\
        \midrule
        \textendash & 66.106 & 2611.501 \\
        \checkmark & \textbf{79.098} & \textbf{3324.573} \\
        \bottomrule
    \end{tabular}
    }
    \caption{\textbf{Ablation study}: Effect of incorporating the pseudo alpha mask. We evaluate the outputs using reference-free, boundary-focused image quality metrics. 
    }
    \vspace{-1.5em}
\label{tab:ablations_alpha}
\end{table}

\section{Conclusion}

We have presented \textbf{HarmoVid}, a novel framework for high-quality video harmonization that effectively addresses the challenges of temporal consistency, identity preservation, and realistic lighting adaptation.  
By introducing a two-stage training strategy—comprising per-frame image harmonization followed by a dedicated Video Deflickering Network—our method is able to generate temporally coherent and visually pleasing harmonized videos even in the absence of paired real-world data. By combining real and synthetic video data, our framework achieves expressive relighting effects while maintaining physical plausibility and temporal smoothness, bridging the gap between controllable synthetic scenarios and complex real-world videos.  

{\noindent\footnotesize\textbf{Acknowledgement.}
This work was conducted during an internship at Adobe Research.
Some of the video assets used in this work are sourced from Adobe Stock and are used in accordance with their licensing terms. These assets are used solely for research and visualization purposes.

This work is partly supported by a National Institute of Health (NIH) project \#R21EB037440.

We thank to Mengwei Ren for sharing the image harmonization model.}
{
    \small
    \bibliographystyle{ieeenat_fullname}
    \bibliography{main}
}


\clearpage
\setcounter{page}{1}
\maketitlesupplementary
\appendix
Along with this supplemental PDF, we provide additional visual materials (e.g., videos) in an website, accessible via \url{https://chedgekorea.github.io/HarmoVid}. We highly recommend readers to refer to the accompanying videos for a comprehensive examination of the visual outcomes.

\section{Overview of Website}
\label{sec:html}
We present the video results corresponding to the figures shown in the main paper. Here, we explain the order and types of results displayed in our website.

\begin{itemize}
    \item We first present video results corresponding to the qualitative comparison in \Cref{fig:main_figure}, showcasing how our \textbf{HarmoVid} performs relative to other image- or video-based harmonization methods.
    \item Next, we show the video results for the deflickering comparison reported in \Cref{tab:deflicker}, which evaluates performance on data processed with per-frame harmonization outputs.
    \item We then provide videos demonstrating the three-stage pipeline in \Cref{tab:ablations_stage}, highlighting the necessity of both Stage 2 and Stage 3.
    \item We also include results for the pseudo-$\alpha$ mask in \Cref{tab:ablations_alpha}, showing that using the pseudo-alpha mask produces smoother boundaries and more realistic harmonization of fine details such as hair.
    \item Finally, we present harmonization results on the DAVIS dataset~\cite{Davis}, illustrating the effectiveness of our method on non-portrait foreground objects.
\end{itemize}

\section{Necessity of Two-Stage Approach}
To train a video harmonization model, paired videos are crucial. Since it is impossible to collect the same motion under different lighting, synthetic data is indispensable, which we generate using Stage 1.

\begin{table}[h!]
    \centering
    \resizebox{1\linewidth}{!}{%
    \begin{tabular}{ccccc}
        \toprule
          \bfseries{Single Stage} & \bfseries{SSIM $\uparrow$} & \bfseries{LPIPS $\downarrow$} & \bfseries{CLIP Score $\uparrow$} & \bfseries{Motion Preservation $\downarrow$} \\
        \midrule

         Real $\rightarrow$ Synthetic & 0.9027 & 0.0783 & 0.9903 & 0.9516 \\
         Real $\leftarrow$ Synthetic & 0.6439 & 0.1258 & 0.9961 & 0.5582 \\
         Real $\leftrightarrow$ Synthetic & 0.9187 & 0.0613 & 0.9911 & 0.9376 \\
        \midrule
        \bfseries{Two Stage} & \textbf{0.9306} & \textbf{0.0554} & \textbf{0.9963} & \textbf{0.5264} \\
        
        \bottomrule
    \end{tabular}
    }
    \caption{Single-Stage Comparison using \Cref{tab:main_comparison} dataset.}
\label{tab:stage}
\end{table}

Training a single-stage on both real video data and flickering synthetic data presents problems. When training with real input and synthetic target, the model successfully applies some harmonization, but temporal consistency is disrupted, resulting in flickering artifacts in the output. Conversely, when training with synthetic input and real target, the model tends to learn that the input should contain flickering. As a result, if an input without flickering is provided during inference, the output closely resembles the input. 

Using a dual-path approach (\Cref{tab:ablations_stage}, Stage 3) allows both types of supervision to be applied simultaneously. However, this does not reduce the domain gap to produce temporally consistent and harmonization; the two paths operate independently. As a result, when using real video input during inference, harmonization occurs to some degree, but the output exhibits flickering (similar to \Cref{tab:stage} 1st row). Therefore, we first apply a deflickering network to reduce flickering in the synthetic data, and use it in Stage 3 to achieve both harmonization and temporal consistency.

\section{Effectiveness of Dual-path Training }
We demonstrate that, during Stage 3 training using the deflickered videos obtained from Stage 2, the dual-path design achieves better performance than the single-path design, as discussed in the main paper.

\begin{table}[h!]
    \centering
    \resizebox{1\linewidth}{!}{%
    \begin{tabular}{ccccc}
        \toprule
          \bfseries{Stage 3} & \bfseries{SSIM $\uparrow$} & \bfseries{LPIPS $\downarrow$} & \bfseries{CLIP Score $\uparrow$} & \bfseries{Motion Preservation $\downarrow$} \\
        \midrule

         Real $\rightarrow$ Synthetic & \underline{0.9287} & \underline{0.0577} & 0.9951 & 0.5510 \\
         Real $\leftarrow$ Synthetic & 0.9202 & 0.0607 & \underline{0.9961} & \underline{0.5271} \\
         Real $\leftrightarrow$ Synthetic & \textbf{0.9306} & \textbf{0.0554} & \textbf{0.9963} & \textbf{0.5264} \\
        
        \bottomrule
    \end{tabular}
    }
    \caption{The real$\rightarrow$synthetic path modifies lighting expressively, while the real$\leftarrow$synthetic path ensures natural harmonization and temporal coherence; together, the model captures both benefits.}
\label{tab:dual_path}
\end{table}

\vspace{-1em}
\section{Computational Cost Analysis}
\vspace{-1em}
\begin{table}[h!]
    \centering
    \resizebox{0.8\linewidth}{!}{%
    \begin{tabular}{cccc}
        \toprule
           & \bfseries{Inference Time (s) $\downarrow$} & \bfseries{GPU (GB) $\downarrow$} \\
        \midrule

         Light-A-Video & 1822 & 42.483 \\
         RelightVid & 199 & 27.234 \\
         \textbf{Ours} & \textbf{168} & \textbf{17.385} \\
        \bottomrule
    \end{tabular}
    }
    \caption{Our method shows better practical efficiency compared to other baselines.}
\label{tab:computational cost}
\end{table}
\vspace{-1em}

\section{User Study Structure Details}
\label{sec:user_study}

\paragraph{Introduction.} 
This study was designed to gather perceptual feedback on three key aspects of video harmonization: overall visual quality, temporal consistency, and identity preservation. By focusing on these aspects, we aim to evaluate not only the realism of individual frames but also the stability and coherence of the foreground object across the video sequence.

\begin{figure}[h]
  \centering
  \includegraphics[width=\linewidth]{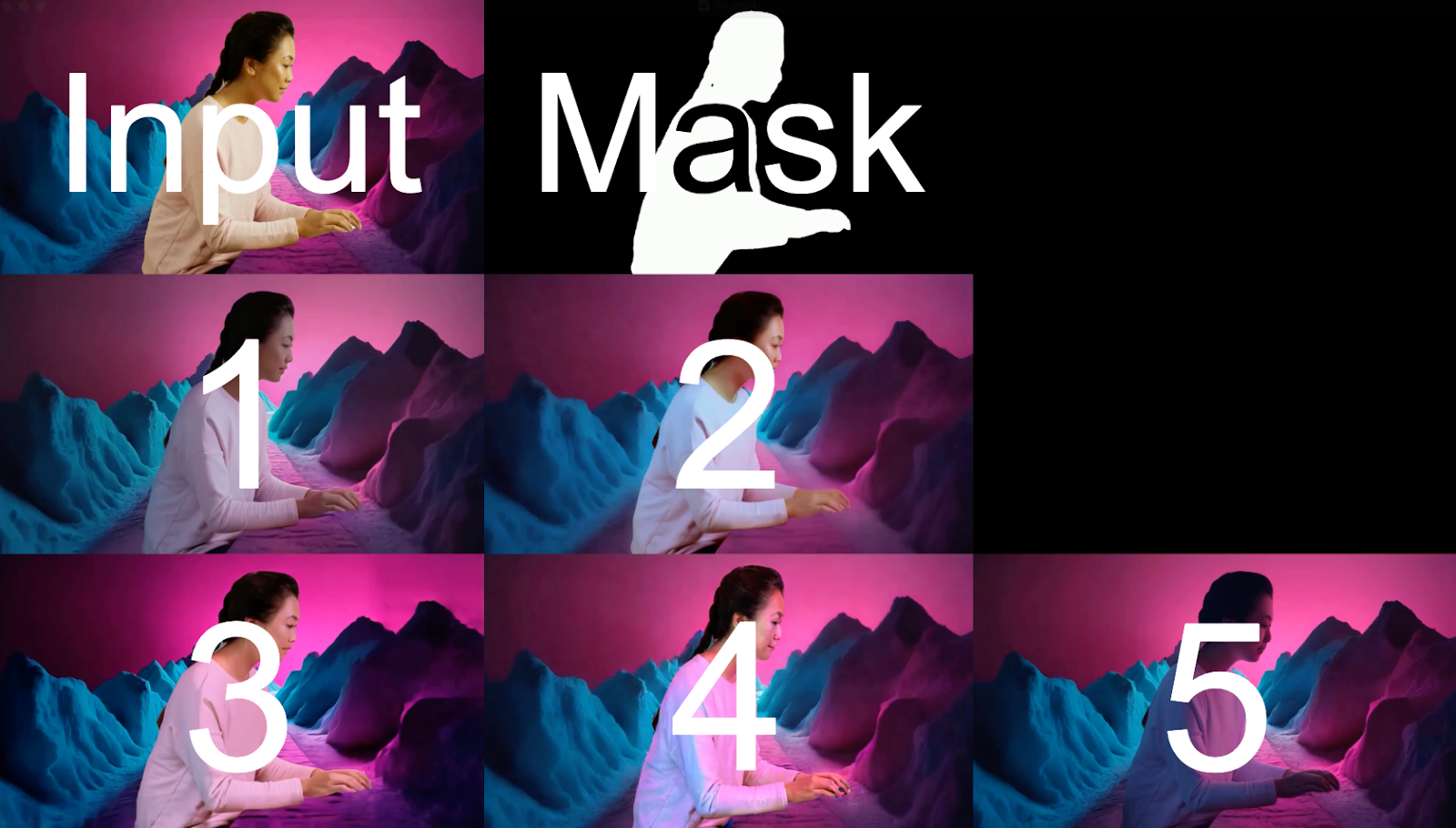}
  \caption{Illustration of an example video from the user study.}
  \label{fig:user_study}
\end{figure}

We then presented explanations for each video example as shown in \Cref{fig:user_study}, where five different results are displayed in random order.
\newpage

\begin{PromptBox}{gray}{Introduction}
In this study, you will be asked to evaluate the visual quality of harmonized videos. Video harmonization aims to make a composited foreground object look naturally integrated with the background scene — for example, by adjusting lighting, color, and shading to match the background lighting.

\noindent{The harmonization process modifies only the white mask region to make it visually consistent with the background scene.}

\noindent{Each video is organized as follows:}

\noindent{the first row contains the input (composited video) and mask (white = foreground, black = background),
the second row contains 2 different harmonized results (1–2),
and the third row contains 3 different harmonized results (3–5).}
\end{PromptBox}
\vspace{-3em}

\paragraph{Section 1: Overall Harmonization Quality.} 
Participants watched 10 harmonized video clips and rated how natural and realistic the harmonized foreground appeared as a whole. They were instructed:

\begin{PromptBox}{gray}{Section 1}
Please ignore small holes or missing areas inside the mask (due to the objects), and focus solely on how natural and consistent the harmonization looks.

\noindent{Please base your ratings only on the harmonization quality, not on the content of the scene or the actions performed.}
\end{PromptBox}

The evaluation question was:

\begin{PromptBox}{gray}{Questions 1-10}
How natural and realistic does the harmonized foreground look in the scene? (in terms of harmonization and relighting)

\noindent{Choose all the realistic videos.}
\end{PromptBox}

This section specifically measures whether the harmonized foreground appears convincingly integrated into the background, reflecting proper lighting, color matching, and shading adjustments.

\paragraph{Section 2: Temporal Consistency and Identity Preservation.} 
The same 10 video clips were evaluated as follows. First they were instructed:

\begin{PromptBox}{gray}{Temporal Consistency \& Identity Preservation}
You will evaluate same 10 harmonized videos based on two aspects:

\noindent{\textbf{Temporal Consistency} — smoothness and stability across frames (i.e., no flickering or unnatural changes).}

\noindent{\textbf{Identity Preservation} — how consistent the subject’s identity and appearance remain throughout the sequence. (how similar the harmonized foreground is to the input foreground.)}
\end{PromptBox}

The evaluation question was:

\begin{PromptBox}{gray}{Questions 11-20}
How temporally consistent is the harmonized foreground across the video?
(Focus on flickering, and smooth transitions between frames)

\noindent{Choose all the temporal consistent videos.}
\end{PromptBox}

This evaluates whether the harmonization is stable over time, which is crucial for video applications, as inconsistent lighting or color shifts can break realism.

\begin{figure}[th]
\begin{PromptBox}{gray}{Questions 21-30}
How consistent does the subject’s (foreground) identity and appearance remain throughout the sequence?

Choose all the videos where the identity is well preserved.
\end{PromptBox}
\end{figure}

This ensures that harmonization does not alter key attributes of the foreground subject, such as facial features, hairstyle, or object shape, which is particularly important for portraits or recognizable objects.

\begin{figure*}[h]
  \centering
  \includegraphics[width=\linewidth]{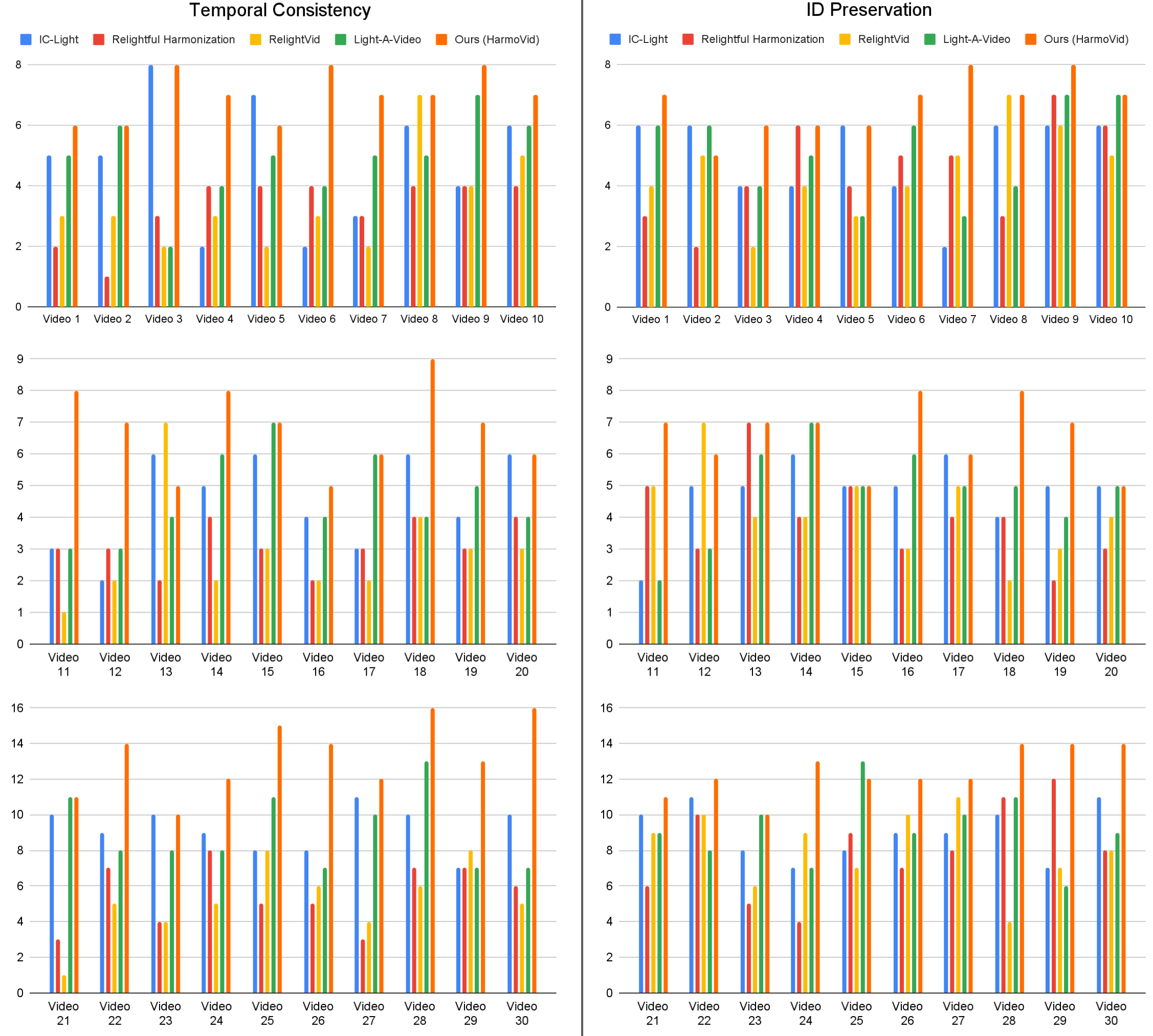}
  \caption{The user study results for temporal consistency and identity preservation.}
  \label{fig:user_study_result2}
\end{figure*}

\begin{figure}[t]
  \centering
  \includegraphics[width=\linewidth]{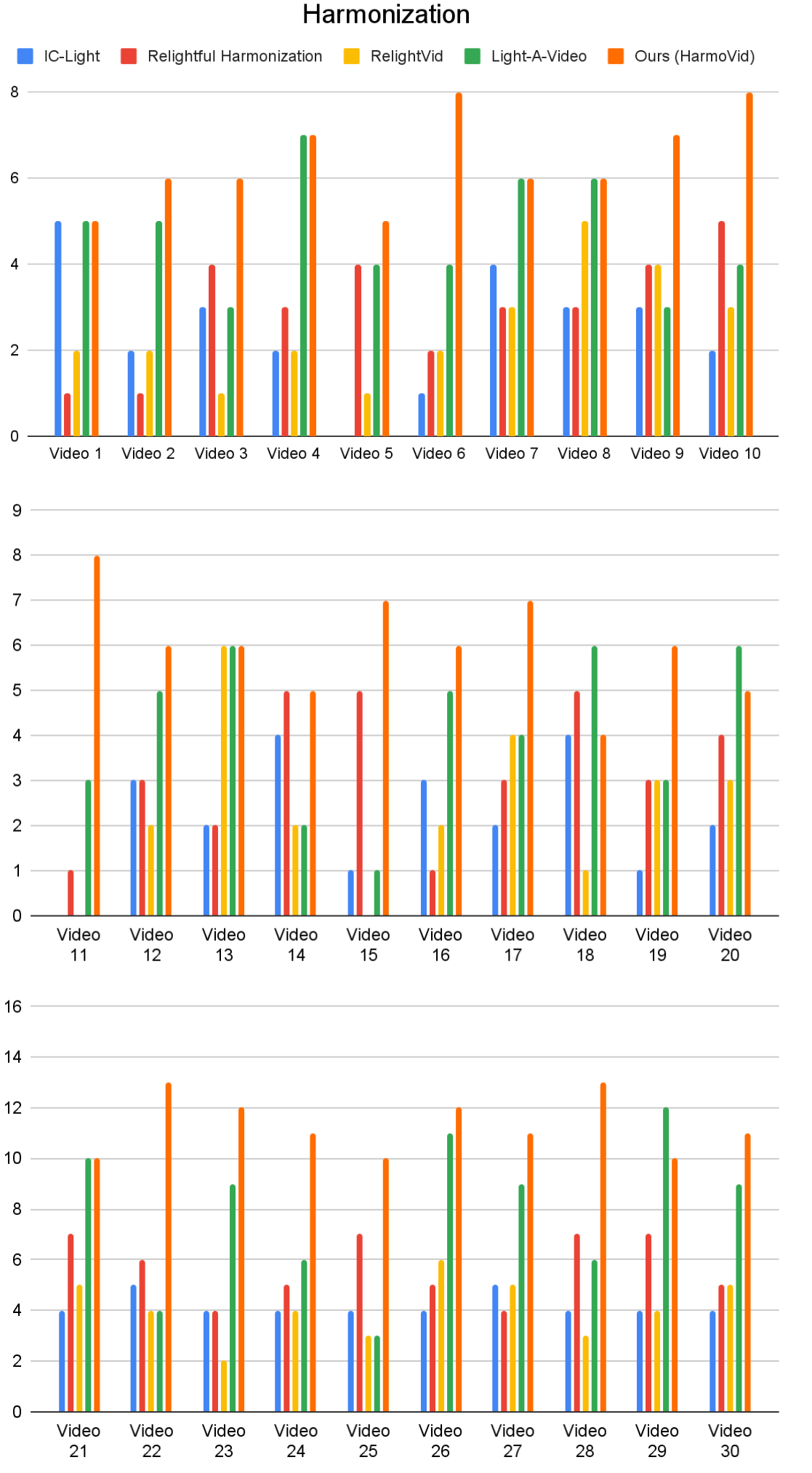}
  \caption{The user study results for overall harmonization quality.}
  \label{fig:user_study_result1}
\end{figure}

\paragraph{Per-Video Results.} In addition to the table results in the main paper (\cref{tab:main_comparison}), we also present per-video results as bar charts in \cref{fig:user_study_result1,fig:user_study_result2}.
The user study was conducted using three different templates, with each row corresponding to one template.
The first, second, and third templates were completed by 8, 9, and 16 participants, respectively.

\end{document}